\documentclass{article}

\usepackage[final]{neurips_2019}

\usepackage[utf8]{inputenc} 
\usepackage[T1]{fontenc}    
\usepackage{hyperref}       
\usepackage{url}            
\usepackage{booktabs}       
\usepackage{amsfonts}       
\usepackage{nicefrac}       
\usepackage{microtype}      

\usepackage{algorithm}
\usepackage{algorithmic}
\usepackage{mathrsfs}

\usepackage{verbatim}
\usepackage{amsmath, amssymb,amsthm, color}
\usepackage{amsthm} 
\usepackage{mathtools}
\usepackage{esvect}
\usepackage{bbm}
\usepackage{dsfont}
\usepackage{tikz}
\usepackage{relsize}
\usepackage{adjustbox}
\usetikzlibrary{arrows,decorations.pathmorphing,fit,positioning}
\usepackage{enumitem}
\usepackage{units}

\usepackage{cleveref}

\usepackage{tikz}

\usepackage{times}
\usepackage{graphicx} 
\usepackage{subcaption}
\usepackage{wrapfig}

\theoremstyle{definition}

\newcommand{\R}{\mathbb{R}}
\newcommand{\abs}[1]{{\left| #1 \right|}}

\newcommand{\spt}[1]{\text{supp}{\left( #1 \right)}}
\newcommand{\WMD}{W\!M\!D}
\newcommand{\RWMD}{RW\!M\!D}

\newcommand{\HOTT}{H\!O\!T\!T}

\long\def\comment#1{}

\title{Hierarchical Optimal Transport\\for Document Representation}

\author{%
  Mikhail Yurochkin$^{1,3}$ \\
  \texttt{mikhail.yurochkin@ibm.com} \\
  \And
  Sebastian Claici$^{2,3}$ \\
  \texttt{sclaici@mit.edu} \\
  \And
  Edward Chien$^{2,3}$ \\
  \texttt{edchien@mit.edu} \\
  \AND
  Farzaneh Mirzazadeh$^{1,3}$ \\
  \texttt{farzaneh@ibm.com} \\
  \And
  Justin Solomon$^{2,3}$ \\
  \texttt{jsolomon@mit.edu} \\
    \AND
  \normalfont{IBM Research,$^1$ MIT CSAIL,$^2$ MIT-IBM Watson AI Lab$^3$}
}

\begin{document}

\maketitle
\begin{abstract}
The ability to measure similarity between documents enables intelligent summarization and analysis of large corpora. Past distances between documents suffer from either an inability to incorporate semantic similarities between words or from scalability issues.
As an alternative, we introduce hierarchical optimal transport as a \emph{meta}-distance between documents, where documents are modeled as distributions over topics, which themselves are modeled as distributions over words. We then solve an optimal transport problem on the smaller topic space to compute a similarity score. 
We give conditions on the topics under which this construction defines a distance, and we relate it to the word mover's distance. 
We evaluate our technique for $k$-NN classification and show better interpretability and scalability with comparable performance to current methods at a fraction of the cost.\footnote{Code: \url{https://github.com/IBM/HOTT}}

\end{abstract}


\section{Introduction}

Topic models like latent Dirichlet allocation (LDA) \citep{blei2003latent} are major workhorses for summarizing document collections. Typically, a topic model represents topics as distributions over the vocabulary (i.e., unique words in the corpus); documents are then modeled as distributions over topics. In this approach, words are vertices of a simplex whose dimension equals the vocabulary size and for which the distance between any pair of words is the same. More recently, word embeddings map words into high-dimensional space such that co-occurring words tend to be closer to each other than unrelated words \citep{mikolov2013distributed, pennington2014glove}. \citet{kusner2015word} combine the geometry of word embedding space with optimal transport to propose the \emph{word mover's distance} (WMD), a powerful document distance metric limited mostly by computational complexity.

As an alternative to WMD, in this paper we combine hierarchical latent structures from topic models with geometry from word embeddings. We propose \emph{hierarchical} optimal topic transport (HOTT) document distances, which combine language information from word embeddings with corpus-specific, semantically-meaningful topic distributions from latent Dirichlet allocation (LDA) \citep{blei2003latent}. This document distance is more efficient and more interpretable than WMD.

We give conditions under which HOTT gives a metric and show how it relates to WMD. We test against existing metrics on $k$-NN classification and show that it outperforms others on average. It performs especially well on corpora with longer documents and is robust to the number of topics and word embedding quality. Additionally, we consider two applications requiring pairwise distances. The first is visualization of the metric with t-SNE \citep{vanDerMaaten2008Visualizing}. The second is link prediction from a citation network, cast as pairwise classification using HOTT features.

\textbf{Contributions.} We introduce \emph{hierarchical} optimal transport to measure dissimilarities between distributions with common structure. We apply our method to document classification, where topics from a topic modeler represent the shared structure. Our approach
\begin{itemize}[leftmargin=*,noitemsep]
    \item is \textbf{computationally efficient}, since HOTT distances involve transport with small numbers of sites;
    \item uses corpus-specific topic and document distributions, providing \textbf{higher-level interpretability};
    \item has \textbf{comparable performance} to WMD and other baselines for $k$-NN classification; and
    \item is \textbf{practical} in applications where all pairwise document distances are needed.
\end{itemize}

\section{Related work}


Document representation and 
similarity assessment 
are key applications 
in learning. 
Many methods 
are based on the bag-of-words (BOW),
which represents documents as vectors in $\R^\abs{V}$, where $\abs{V}$ is the vocabulary size; each coordinate equals the number of times a word appears. Other weightings include term frequency inverse document frequency (TF-IDF)~\citep{luhn1957statistical,sparck1972statistical} and 
latent semantic indexing (LSI)~\citep{deerwester1990indexing}. Latent Dirichlet allocation (LDA) \citep{blei2003latent} is a hierarchical Bayesian model where documents are represented as admixtures of latent topics and admixture weights provide low-dimensional representations. These representations equipped with the $l_2$ metric 
comprise early examples of document dissimilarity scores.

Recent document distances employ more sophisticated methods. WMD incorporates word embeddings to account for word similarities~\citep{kusner2015word} (see \S\ref{sec:background}). \citet{huang2016supervised} extend WMD to the supervised setting, modifying  
embeddings so that documents in the same class are close and documents from different classes are far. Due to computational complexity, these approaches are impractical for large corpora or documents with many unique words.

\citet{wu2017topic} attempt to address the complexity of WMD via a topic mover's distance (TMD). While their $k$-NN classification results are comparable to WMD, they use significantly more topics, generated with a Poisson infinite relational model. This reduces semantic content and interpretability, with less significant computational speedup. They also do not leverage language information from word embeddings or otherwise. \citet{xu2018distilled} jointly learn topics and word embeddings, limiting the complexity to 
under a hundred words, which is not suited for natural language processing.  

\citet{wu2018word} approximate WMD using a random feature kernel. In their method, the WMD from corpus documents to a selection of random short documents facilitates approximation of pairwise WMD. The resulting word mover's embedding (WME) has similar performance with significant speedups. Their method, however, requires parameter tuning in selecting the random document set 
and lacks topic-level interpretability. Additionally, they do not show full-metric applications. Lastly, \citet{wan2007document}, whose work predates \citep{kusner2015word}, applies transport to blocks of text.




\section{Background}
\label{sec:background}

\textbf{Discrete optimal transport.}
Optimal transport (OT) is a rich theory; we only need a small part and refer the reader to \citep{villani_optimal_2009,santambrogio_optimal_2015} for mathematical foundations and to \citep{peyre2018computational,solomon2018optimal} for applications. 
%
Here, we focus on discrete-to-discrete OT. 

Let $\mathbf{x} = \{x_1, \ldots, x_n\}$ and $\mathbf{y} = \{y_1, \ldots, y_m\}$ be two sets of points (sites) in a metric space. Let $\Delta^n \subset \R^{n+1}$ denote the probability simplex on $n$ elements, and let $p \in \Delta^n$ and $q \in \Delta^m$ be distributions over $\mathbf{x}$ and $\mathbf{y}$. 
Then, the $1$-Wasserstein distance between $p$ and $q$ is  
\begin{equation}
     \label{eq:OTopt}
     W_1(p,q) = \left\{
     \begin{array}{rl}
     \min_{\Gamma\in\R^{n\times m}_+}\ & \sum_{i,j} C_{i,j}\Gamma_{i,j}\\
     \textrm{subject to}\ & \sum_j \Gamma_{i, j} = p_i\textrm{ and }
     \sum_i \Gamma_{i, j} = q_j,
     \end{array}
     \right.
\end{equation}
where the cost matrix $C$ has entries $C_{i,j}=d(x_i,y_j)$, where $d(\cdot,\cdot)$ denotes the distance. The constraints allow $\Gamma$ to be interpreted as a transport plan or matching between $p$ and $q$. 
%
The linear program~\eqref{eq:OTopt} can be solved using the Hungarian algorithm~\citep{kuhn1955hungarian}, with complexity $O(l^3 \log l)$ where $l = \max(n,m)$. While entropic regularization can accelerate OT in learning environments \citep{cuturi2013sinkhorn}, it is most successful when the support of the distributions is large as it has complexity $O(l^2 / \varepsilon^2)$. In our case, the number of topics in each document is small, and the linear program is typically faster if we need an accurate solution (i.e. if $\varepsilon$ is small). 


%

\textbf{Word mover's distance.} Given an embedding of a vocabulary as $V \subset \R^n$, the Euclidean metric puts a geometry on the words in $V$. A corpus $D = \{d^1, d^2, \ldots d^{\abs{D}}\}$ can be represented using distributions over $V$ via a normalized BOW. In particular, $d^i \in \Delta^{l_i}$, where $l_i$ is the number of unique words in a document $d^i$, and
\(d^i_j = \nicefrac{c^i_j}{\abs{d^i}}, \)
where $c^i_j$ is the count of word $v_j$ in $d^i$ and $\abs{d^i}$ is the number of words in $d^i$. The WMD between documents $d^1$ and $d^2$ is then
$ \WMD(d^1,d^2) = W_1 (d^1,d^2). $

The complexity of computing WMD depends heavily on $l = \max(l_1,l_2)$; for longer documents, $l$ may be a significant fraction of $\abs{V}$. To evaluate the full metric on a corpus, the complexity is $O(\abs{D}^2 l^3 \log l)$, since $\WMD$ must be computed pairwise. \citet{kusner2015word} test WMD for $k$-NN classification. To circumvent complexity issues, they introduce a pruning procedure using a relaxed word mover's distance (RWMD) to lower-bound WMD. On the larger \textsc{20news} dataset, they additionally remove infrequent words by using only the top 500 words to generate a representation.

\section{Hierarchical optimal transport}
\label{sec:tmd}

Assume a topic model produces corpus-specific topics $T = \{t_1, t_2, \ldots, t_{\abs{T}}\} \subset \Delta^{\abs{V}}$, which are distributions over words, as well as document distributions $\bar{d}^i \in \Delta^{\abs{T}}$ over topics. WMD defines a metric $\WMD(t_i,t_j)$ between topics; we consider discrete transport over $T$ as a metric space.

We define the hierarchical topic transport distance (HOTT) between documents $d^1$ and $d^2$ as
\[ \HOTT(d^1,d^2) = W_1\left(\sum_{k=1}^{\abs{T}} \bar{d}^1_k \delta_{t_k}, \sum_{k=1}^{\abs{T}} \bar{d}^2_k \delta_{t_k}\right), \]
where each Dirac delta $\delta_{t_k}$ is a probability distribution supported on the corresponding topic $t_k$ and where the ground metric is WMD between topics as distributions over words.
The resulting transport problem leverages topic correspondences provided by WMD in the base metric. This explains the \emph{hierarchical} nature of our proposed distance.


Our construction uses transport \emph{twice}:  WMD provides topic distances, which are subsequently the costs in the HOTT problem.  This hierarchical structure greatly reduces runtime, since $\abs{T} \ll l$; the costs for HOTT can be precomputed once per corpus. The expense of evaluating pairwise distances is drastically lower, since pairwise distances between topics may be precomputed and stored. Even as document length and corpus size increase, the transport problem for HOTT remains the same size. Hence, full metric computations are feasible on larger datasets with longer documents.

When computing $\WMD(t_i,t_j)$, we reduce computational time by truncating topics to a small amount of words carrying the majority of the topic mass and re-normalize. This procedure is motivated by interpretability considerations and estimation variance of the tail probabilities. On the interpretability side, LDA topics are often displayed using a few dozen top words, providing a human-understandable tag. Semantic coherence, a popular topic modeling evaluation metric, also is based on heavily-weighted words and was demonstrated to align with human evaluation of topic models 
\citep{newman2010automatic}. Moreover, any topic modeling inference procedure, e.g.\ Gibbs sampling \citep{griffiths2004finding}, has estimation variance that may dominate tail probabilities, making them unreliable. Hence, we truncate 
to the top 20 words when computing WMD between topics. We empirically verify that truncation to any small number of words performs equally well in \S\ref{sec:sensitivity}.

\begin{figure*}[t]\centering
\includegraphics[width=\linewidth]{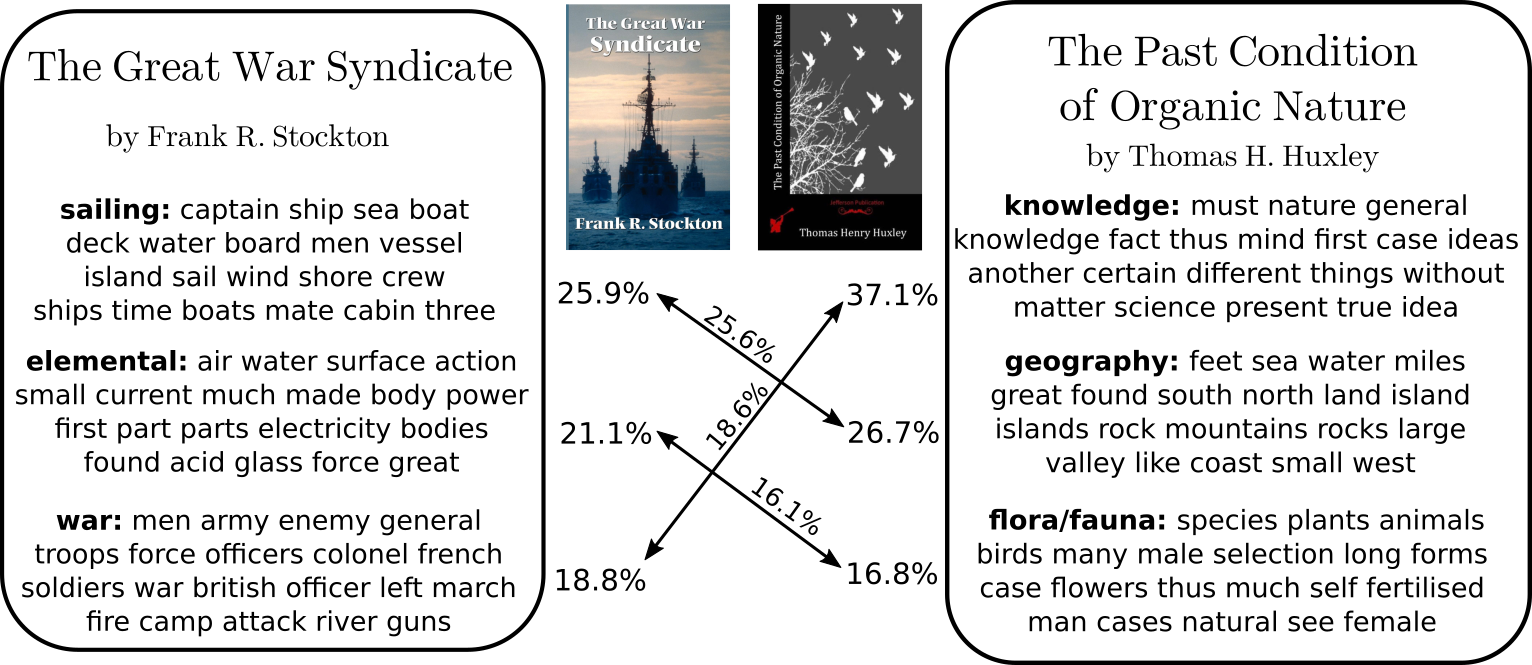}
\caption{Topic transport interpretability. We show two books from \textsc{Gutenberg} and their heaviest-weighted topics (bolded topic names are manually assigned). The first involves steamship warfare, while the second involves 
biology. Left and right column percentages indicate the weights of the topics in the corresponding texts. Percentages labeling the arrows indicate the transported mass between the corresponding topics, which match semantically-similar topics.
}
\label{fig:interp}
\end{figure*}

In topic models, documents are assumed to be represented by a small subset of topics of size $\kappa_i \ll \abs{T}$ (e.g., in Figure \ref{fig:interp}, \emph{books} are majorly described by three topics), but in practice document topic proportions tend to be dense with little mass outside of the dominant topics. \citet{williamson2010ibp} propose an LDA extension enforcing sparsity of the topic proportions, at the cost of slower inference. When computing HOTT, we simply truncate LDA topic proportions at $1/(\abs{T}+1)$, the value below LDA's uniform topic proportion prior, and re-normalize.  This reduces complexity of our approach without performance loss as we show empirically in \S\ref{sec:knn} and \S\ref{sec:sensitivity}.

\textbf{Metric properties of HOTT.} If each document can be uniquely represented as a linear combination of topics $d^i = \sum_{k=1}^{|T|}\bar{d}_k^i t_k$, and each topic is unique, then $\HOTT$ is a metric on document space. We present a brief proof in Appendix \ref{supp:metric}.

\textbf{Topic-level interpretability.} The additional level of abstraction promotes higher-level interpretability at the level of topics as opposed to dense word-level correspondences from WMD.
We provide an example in Figure~\ref{fig:interp}. This diagram illustrates two books from the \textsc{Gutenberg} dataset and the semantically meaningful transport between their three most heavily-weighted topics. Remaining topics and less prominent transport terms account for the remainder of the transport plan not illustrated.

\textbf{Relation to WMD.} \label{sec:TMDbound}
First we note that if $\abs{T} = \abs{V}$ and topics consist of single words covering the vocabulary, then HOTT becomes WMD. In well-behaved topic models, this is expected as $\abs{T} \to \abs{V}$. Allowing $\abs{T}$ to vary produces different levels of granularity for our topics as well as a trade-off between computational speed and topic specificity. When $\abs{T} \ll \abs{V}$, we argue that WMD is upper bounded by HOTT and two terms that represent topic modeling loss. By the triangle inequality, 
\begin{equation}
    \WMD(d^i, d^j) \leq\!W_1\!\left(d^i,\!\sum_{k=1}^{|T|}\!\bar{d}^i_k t_k\!\right) 
    \!+\!W_1\!\left(\!\sum_{k=1}^{|T|}\!\bar{d}^i_k t_k,\!\sum_{k=1}^{|T|}\!\bar{d}^j_k t_k\!\right)\!+\!
    W_1\!\left(\!\sum_{k=1}^{|T|}\!\bar{d}^j_k t_k, d^j\!\right)\!.
\end{equation}

LDA inference minimizes 
$\mathrm{KL}\!(d^i\| \sum_{k=1}^{|T|} \bar{d}^i_k t_k)$ over topic proportions $\bar{d^{i}}$ for a given document $d^i$; hence, we look to relate Kullback--Leibler divergence to $W_1$. In finite-diameter metric spaces, $W_1(\mu, \nu) \leq \text{diam}(X) \sqrt{\frac{1}{2}\text{KL}(\mu \| \nu)}$, which follows from inequalities relating Wasserstein distances to KL divergence \citep{otto2000generalization}.  
The middle term satisfies the following inequality:
\begin{equation}
    W_1\!\left(\sum_{k=1}^{|T|} \bar{d}^i_k t_k, \sum_{k=1}^{|T|} \bar{d}^j_k t_k\right)
    \leq W_1\!\left(\sum_{k=1}^{|T|} \bar{d}^i_k \delta_{t_k}, \sum_{k=1}^{|T|} \bar{d}^j_k \delta_{t_k}\right),
\end{equation}
where on the right we have $\HOTT(d^1,d^2)$. 
The optimal topic transport on the right implies an equal-cost transport of the corresponding linear combinations of topic distributions on the left. The inequality follows since $W_1$ gives the \emph{optimal} transport cost. Combining into a single inequality, 
\begin{equation*}
    \WMD(d^i, d^j) \leq \HOTT(d^i, d^j)+\text{diam}(X)\left[ \sqrt{\frac{1}{2} \text{KL}\left(d^j\middle\| \sum_{k=1}^{|T|} \bar{d}^j_k t_k\right)} + \sqrt{\frac{1}{2} \text{KL}\left(d^i\middle\| \sum_{k=1}^{|T|} \bar{d}^i_k t_k\right)} \right].
\end{equation*}

WMD involves a large transport problem and \citet{kusner2015word} propose relaxed WMD (RWMD), a relaxation via a lower bound (see also \citet{atasu2019linear} for a GPU-accelerated variant). We next show that RWMD is not always a good lower bound on WMD.


\textbf{RWMD--Hausdorff bound.} \label{sec:RWMDbound} Consider the optimization in \eqref{eq:OTopt} for calculating $\WMD(d^1,d^2)$, and remove the marginal constraint on $d^2$. The resulting optimal $\Gamma$ is no longer a transport plan, but rather moves mass on words in $d^1$ to their nearest words in $d^2$, only considering the support of $d^2$ and not its density values. Removing the marginal constraint on $d^1$ produces symmetric behavior; $\RWMD(d^1,d^2)$ is defined to be the larger cost of these relaxed problems. 

Suppose that word $v_j$ is shared by $d^1$ and $d^2$. Then, the mass on $v_j$ in $d^1$ and $d^2$ in each relaxed problems will not move and contributes zero cost. In the worst case, if $d^1$ and $d^2$ contain the same words, i.e., $\spt{d^1} = \spt{d^2}$, then $\RWMD(d^1,d^2) = 0$. More generally, the closer the supports of two documents (over $V$), the looser RWMD might be as a lower bound.  
\begin{wrapfigure}[11]{r}{.27\linewidth}\centering
\begin{tabular}{cc}
\includegraphics[width=\linewidth]{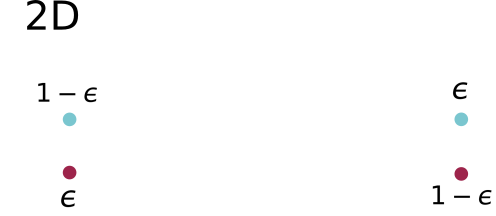}\\\hline
\includegraphics[width=\linewidth]{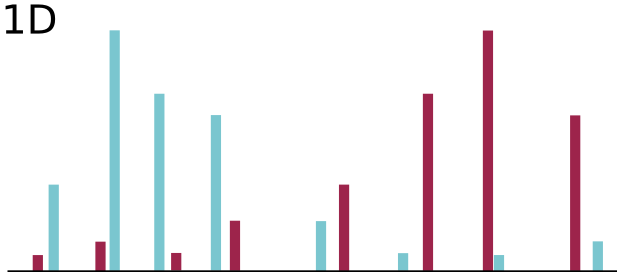}
\end{tabular}
\vspace{-.05in}
\caption{RWMD as a poor approximation to WMD }
\label{fig:RWMDfails}
\end{wrapfigure}

Figure \ref{fig:RWMDfails} illustrates two examples.  In the 2D example, $1-\epsilon$ and $\epsilon$ denote the masses in the teal and maroon documents. The 1D example uses histograms to represent masses in the two documents. In both, RWMD is nearly zero as masses do not have far to move, while the WMD will be larger thanks to the constraints.

To make this precise we provide the following tight upper bound: $\RWMD(d^1,d^2) \leq d_H\!(\spt{d^1},\spt{d^2})$, the Hausdorff distance between the supports of $d^1$ and $d^2$. Let $X = \spt{d^1}$ and $Y = \spt{d^2}$; and let $\RWMD_1(d^1,d^2)$ and $\RWMD_2(d^1,d^2)$ denote the relaxed optimal values when the marginal constraints on $d^1$ and $d^2$ are kept, respectively:
\begin{align*}
    d_H(X, Y) &= \max\left(\sup_{x\in X}\inf_{y\in Y} d(x, y), \sup_{y\in Y}\inf_{x\in X} d(x, y)\right) \\
    & \geq \max\left(\RWMD_1(d^1,d^2),\RWMD_2(d^1,d^2) \right) = \RWMD(d^1,d^2).
\end{align*}
The inequality follows since the left argument of the max is the furthest mass must travel in the solution to $\RWMD_1$, while the right is the furthest mass must travel in the solution to $\RWMD_2$. It is tight if the documents have singleton support and whenever $d^1$ and $d^2$ are supported on parallel affine subspaces and are translates in a normal direction. A 2D example is in Figure \ref{fig:RWMDfails}. 

The preceding discussion suggests that RWMD is not an appropriate way to speed up WMD for long documents with overlapping support, scenario where WMD computational complexity is especially prohibitive. The \textsc{Gutenberg} dataset showcases this failure, in which documents frequently have common words. Our proposed HOTT does not suffer from this failure mode, while being significantly faster and as accurate as WMD. We verify this in the subsequent experimental studies. In the Appendix \ref{supp:rwmd_relation} we present a brief empirical analysis relating HOTT and RWMD to WMD in terms of Mantel correlation and a Frobenius norm.

\section{Experiments}
\label{sec:experiments}
We present timings for metric computation and consider applications where distance between documents plays a crucial role: $k$-NN classification, low-dimensional visualization, and link prediction.

\subsection{Computational timings}

\begin{figure*}[t]\centering
\includegraphics[width=\textwidth]{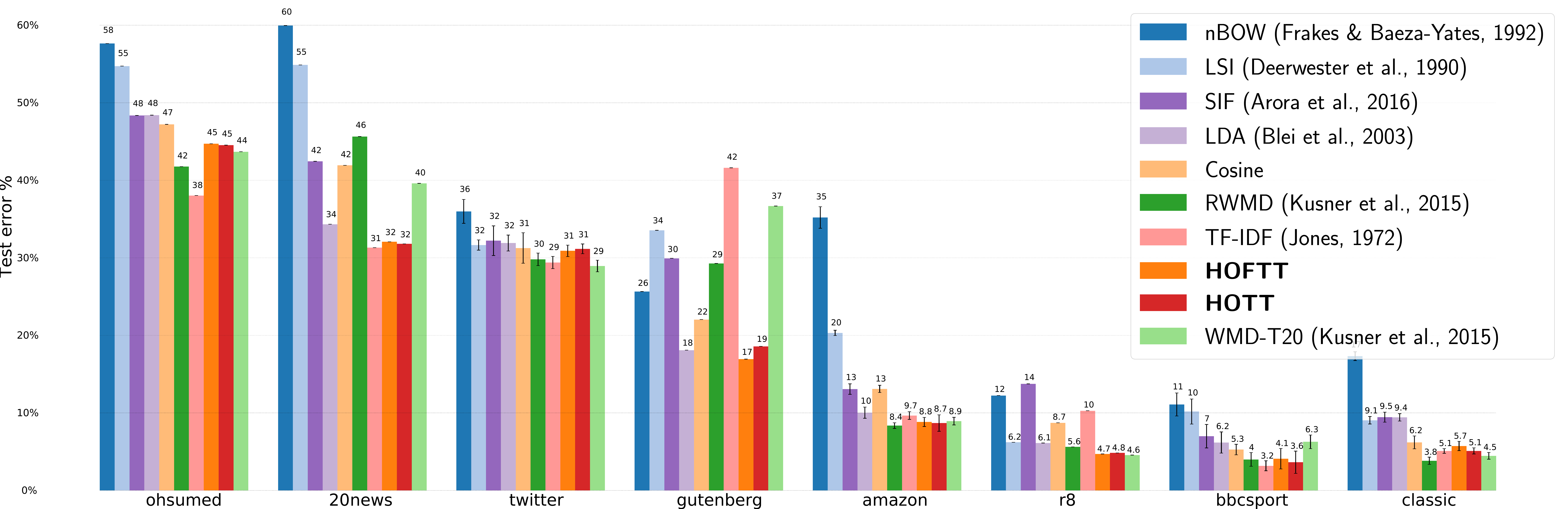}
\caption{$k$-NN classification performance across datasets}
\label{fig:kNN}
\end{figure*}

\textbf{HOTT implementation.} During training, we fit LDA with 70 topics using a Gibbs sampler \citep{griffiths2004finding}. Topics are truncated to the 20 most heavily-weighted words and renormalized. The pairwise distances between topics $\WMD(t_i,t_j)$ are precomputed with words embedded in $\R^{300}$ using \textit{GloVe} \citep{pennington2014glove}. To evaluate HOTT at testing time, a few iterations of the Gibbs sampler are run to obtain topic proportions $\bar{d}^i$ of a new document $d^i$. When computing HOTT between a pair of documents we truncate topic proportions at $1/(\abs{T}+1)$ and renormalize. Every instance of the OT linear program is solved using Gurobi \citep{gurobi}.

We note that LDA inference may be carried out using any other approaches, e.g. stochastic/streaming variational inference \citep{hoffman2013stochastic, broderick2013streaming} or geometric algorithms \citep{yurochkin2016geometric,yurochkin2019dirichlet}. We chose the MCMC variant \citep{griffiths2004finding} for its strong theoretical guarantees, simplicity and wide adoption in the topic modeling literature.

    
    
\textbf{Topic computations.}
 The preprocessing steps of our method---computing LDA topics and the topic to topic pairwise distance matrix---are dwarfed by the cost of computing the full document-to-document pairwise distance matrix. The complexity of base metric computation in our implementation is $O(\abs{T}^2)$, since $\abs{\spt{t_i}} = 20$ for all topics, leading to a relatively small OT instance.

\textbf{HOTT computations.} 
All distance computations were implemented in Python 3.7 and run on an Intel i7-6700K at 4GHz with 32GB of RAM. 
Timings for pairwise distance computations are in Table \ref{table:datasets_speed} (right). HOTT outperforms RWMD and WMD in terms of speed as it solves a significantly smaller linear program. On the left side of Table \ref{table:datasets_speed} we summarize relevant dataset statistics: $\abs{D}$ is the number of documents; $\abs{V}$ is the vocabulary size; intersection over union (IOU) characterizes average overlap in words between pairs of documents; {\sc{avg}$(l)$} is the average number of unique words per document and {\sc{avg}$(\kappa)$} is the average number of major topics (i.e., after truncation) per document.

\subsection{$k$-NN classification}
\label{sec:knn}

We follow the setup of \citet{kusner2015word} to evaluate performance of HOTT on $k$-NN classification.

\textbf{Datasets.} We consider 8 document classification datasets: BBC sports news articles (\textsc{bbcsport}) labeled by sport; tweets labeled by sentiments (\textsc{twitter}) \citep{sanders2011sanders}; Amazon reviews labeled by category (\textsc{amazon}); Reuters news articles labeled by topic (\textsc{reuters}) (we use the 8-class version and train-test split of \citet{cachopo2007improving}); medical abstracts labeled by cardiovascular disease types (\textsc{ohsumed}) (using 10 classes and train-test split as in \citet{kusner2015word}); sentences from scientific articles labeled by publisher (\textsc{classic}); newsgroup posts labeled by category (\textsc{20news}), with ``by-date'' train-test split and removing headers, footers and quotes;\footnote{\url{https://scikit-learn.org/0.19/datasets/twenty_newsgroups.html}} and Project Gutenberg full-length books from 142 authors (\textsc{gutenberg}) using the author names as classes and 80/20 train-test split in the order of document appearance. For \textsc{gutenberg}, we reduced the vocabulary to the most common 15000 words. For \textsc{20news}, we removed words appearing in $\leq\!5$ documents.

\begin{table}[t]
\begin{center}
\begin{small}\scriptsize
\begin{sc}
\caption{Dataset statistics and document pairs per second; higher is better. HOTT has higher throughput 
and excels on long documents with large portions of the vocabulary (as in \textsc{gutenberg}).
}
\label{table:datasets_speed}
\vskip 0.1in
\begin{tabular}{l|r@{\quad}r@{\quad}r@{\quad}r@{\quad}r@{\quad}r|c@{\quad}c@{\quad}c@{\quad}c@{\quad}c}
\toprule
 & \multicolumn{6}{c|}{Dataset statistics} & \multicolumn{5}{c}{Pairs per second} \\
\midrule
Dataset & $\abs{D}$ & $\abs{V}$ & IOU & avg$(l)$ & avg$(\kappa)$ & Classes & RWMD & WMD & WMDT20 & HOFTT & HOTT\\
\midrule
bbcsport   & 737 & 3657 & 0.066 & 116.5 & 11.7 & 5 & 1494 & 526 & 1545 & 2016 & \textbf{2548}\\
twitter & 3108 & 1205 & 0.029 & 9.7 & 6.3 & 3 & \textbf{2664} & 2536 & 2194 & 1384 & 1552\\
ohsumed    & 9152 & 8261 & 0.046 & 59.4 & 11.0 & 10 & 454 & 377 & 473 & 829 & \textbf{908}\\
classic    & 7093 & 5813 & 0.017 & 38.5 & 8.7 & 4 & 816 & 689 & 720 & 980 & \textbf{1053}\\
reuters8     & 7674 & 5495 & 0.06 & 35.7 & 8.7 & 8 & 834 & 685 & 672 & 918 & \textbf{989}\\
amazon     & 8000 & 16753 & 0.019 & 44.3 & 9.0 & 4 & 289 & 259 & 253 & 927 & \textbf{966}\\
20news      & 13277 &  9251 & 0.011 & 69.3 & 10.5 & 20 & 338 & 260 & 384 & 652 & \textbf{699}\\
gutenberg   & 3037 & 15000 & 0.25 & 4367 & 13.3 & 142 & 2 & 0.3 & 359 & 1503 & \textbf{1720}\\
\bottomrule
\end{tabular}
\end{sc}
\end{small}
\end{center}
\vspace{.1in}
\end{table}


\textbf{Baselines.}
We focus on evaluating HOTT and a variation without topic proportion truncation (HOFTT: hierarchical optimal full topic transport) as alternatives to RWMD in a variety of metric-dependent tasks. As demonstrated by the authors, RWMD has nearly identical performance to WMD, while being more computationally feasible. Additionally, we analyze a na\"ive approach for speeding-up WMD where we truncate documents to their top 20 unique words (WMD-T20), making complexity comparable to HOTT (yet $20>${\sc{avg}}$(\kappa)$ on all datasets). For $k$-NN classification, we also consider baselines that represent documents in vector form and use Euclidean distances: normalized bag-of-words (nBOW) \citep{frakes1992information}; latent semantic indexing (LSI) \citep{deerwester1990indexing}; latent Dirichlet allocation (LDA) \citep{blei2003latent} trained with a Gibbs sampler \citep{griffiths2004finding}; and term frequency inverse document frequency (TF-IDF) \citep{sparck1972statistical}. We omit comparison to embedding via BOW weighted averaging as it was shown to be inferior to RWMD by \citet{kusner2015word} (i.e., Word Centroid Distance) and instead consider smooth inverse frequency (SIF), a recent document embedding method by \citet{arora2016simple}. We also compare to bag-of-words, where neighbors are identified using cosine similarity (Cosine).  We use same pre-trained \textit{GloVe} embeddings for HOTT, RWMD, SIF and truncated WMD and set the same number of topics $\abs{T}=70$ for HOTT, LDA and LSI; we provide experiments testing parameter sensitivity.

\begin{wrapfigure}[15]{r}{.4\linewidth}
\vspace{-.1in}
\includegraphics[width=\linewidth]{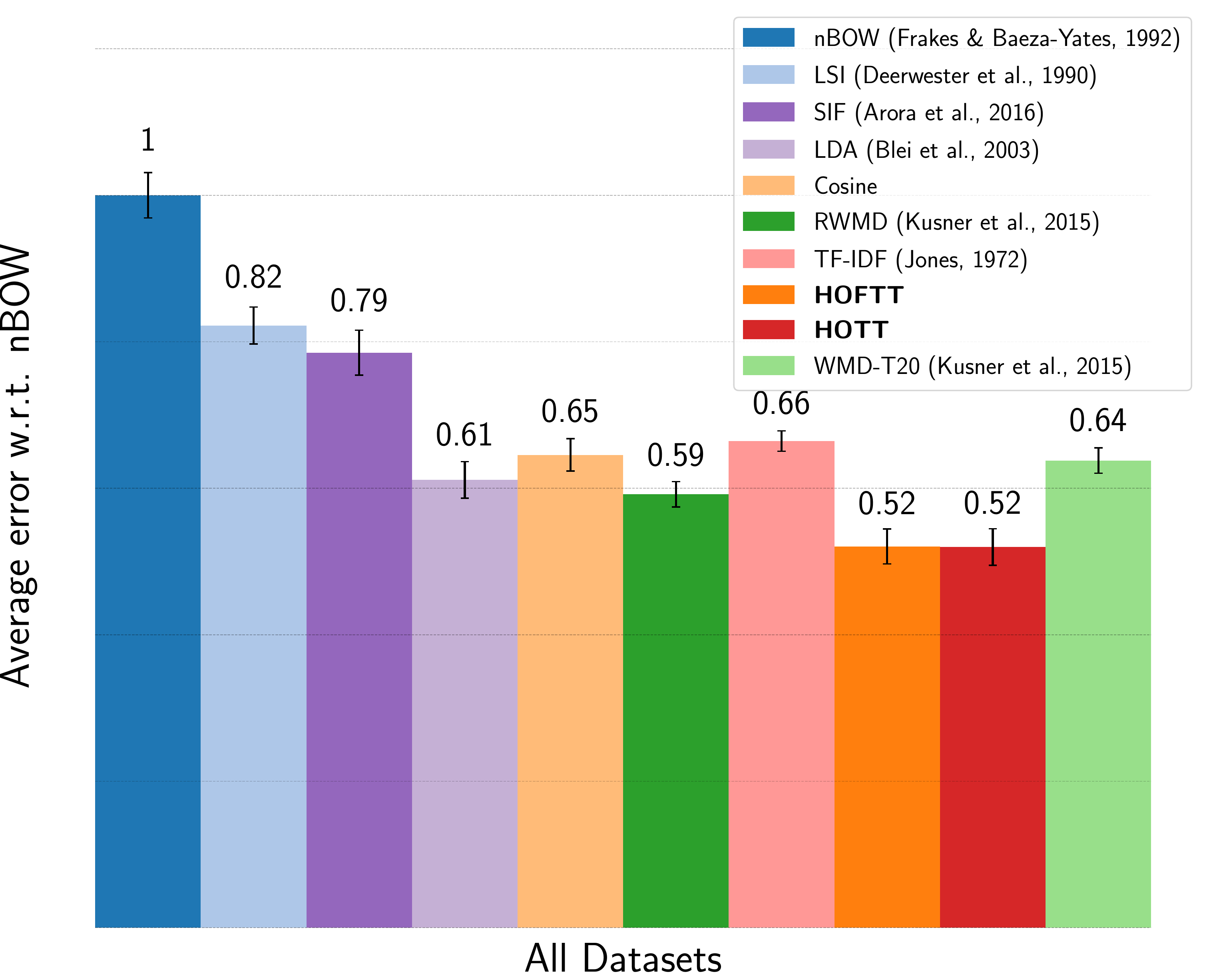}
\caption{Aggregated $k$-NN classification performance normalized by nBOW}
\label{fig:overallkNN}
\end{wrapfigure}

\textbf{Results.} 
We evaluate each method on $k$-NN classification (Fig.\ \ref{fig:kNN}). 
There is no uniformly best method, but HOTT performs best on average (Fig.\ \ref{fig:overallkNN})
We highlight the performance on the \textsc{Gutenberg} dataset compared to RWMD. 
We anticipate poor performance of RWMD on \textsc{Gutenberg}, since books contain more words, which can make RWMD degenerate (see \S\ref{sec:RWMDbound} and Fig.\ \ref{fig:RWMDfails}). 
Also note strong performance of TF-IDF on \textsc{ohsumed} and \textsc{20news}, which differs from results of \citet{kusner2015word}. We believe this is due to a different normalization scheme. We used \textit{TfidfTransformer} from scikit-learn \citep{pedregosa2011scikit} with default settings. We conclude that HOTT is most powerful, both computationally (Table \ref{table:datasets_speed} right) and as a distance metric for $k$-NN classification (\Cref{fig:kNN,fig:overallkNN}), on larger corpora of longer documents, whereas on shorter documents both RWMD and HOTT perform similarly.

Another interesting observation is the effect of truncation: HOTT performs as well as HOFTT, meaning that truncating topic proportions of LDA does not prevent us from obtaining high-quality document distances in less computational time, whereas truncating unique words for WMD degrades its performance. This observation emphasizes the challenge of speeding up WMD, i.e. WMD \emph{cannot} be made computationally efficient using truncation without degrading its performance. WMD-T20 is slower than HOTT (Table \ref{table:datasets_speed}) and performs noticeably worse (Figure \ref{fig:overallkNN}). Truncating WMD further will make its performance even worse, while truncating less will quickly lead to impractical run-time.

In Appendix \ref{supp:experiments}, we complement our results considering 2-Wasserstein distance, and stemming, a popular text pre-processing procedure for topic models to reduce vocabulary size. HOTT continues to produce best performance on average. We restate that in all main text experiments we used 1-Wasserstein (i.e. eq. \eqref{eq:OTopt}) and did not stem, following experimental setup of \citet{kusner2015word}.

\subsection{Sensitivity analysis of HOTT}
\label{sec:sensitivity}
We analyze senstitivity of HOTT with respect to its components: word embeddings, number of LDA topics, and topic truncation level.

\begin{figure*}[t]\centering
\begin{subfigure}{.3\textwidth}
  \centering
  \captionsetup{justification=centering}
\includegraphics[width=\linewidth]{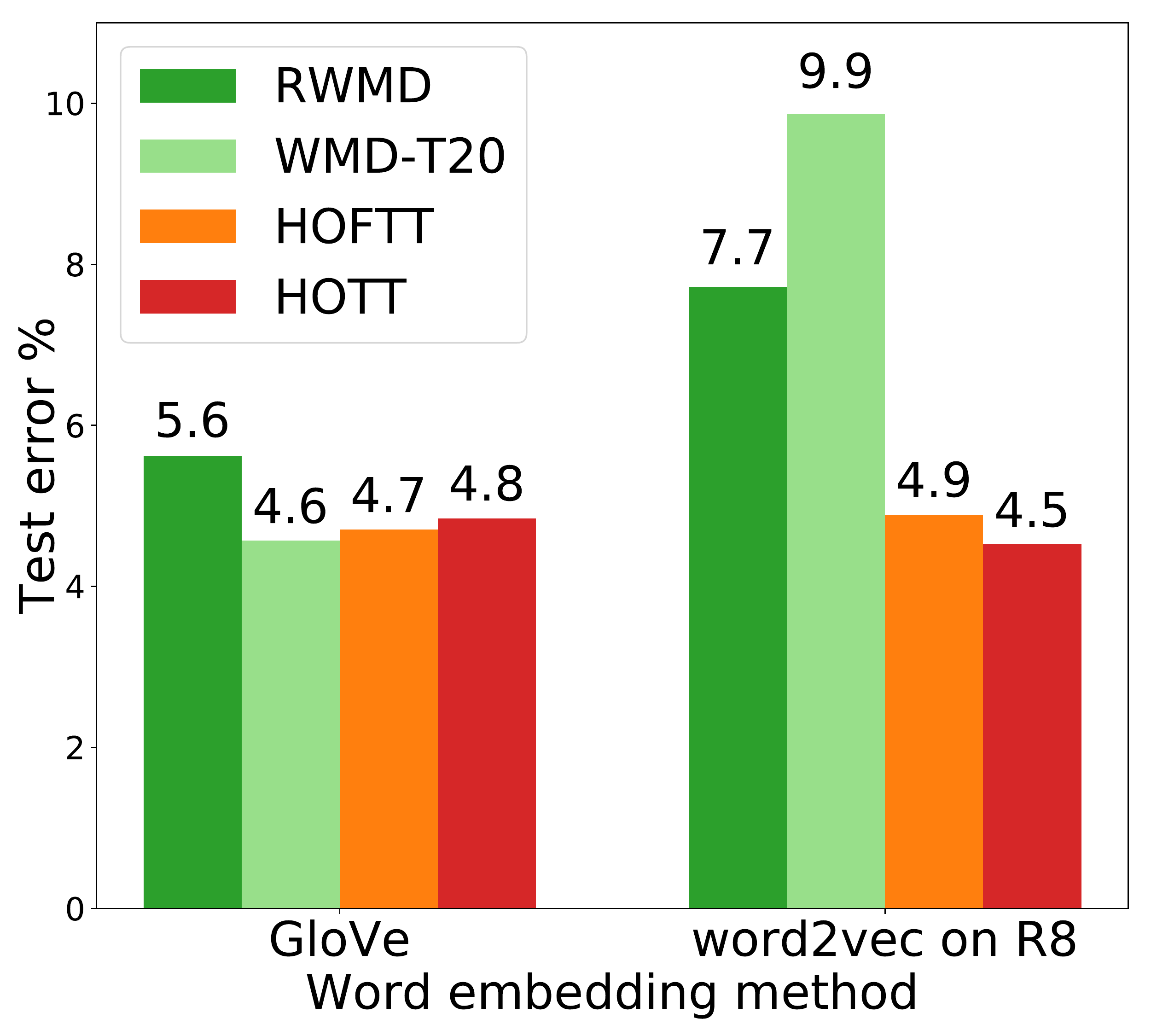}
\vspace{-0.24in}
\caption{Embedding sensitivity on \textsc{reuters}}
\label{fig:r8-embed}
\end{subfigure}
\begin{subfigure}{.33\textwidth}
  \centering
  \captionsetup{justification=centering}
\vspace{-.14in}
\includegraphics[width=\linewidth]{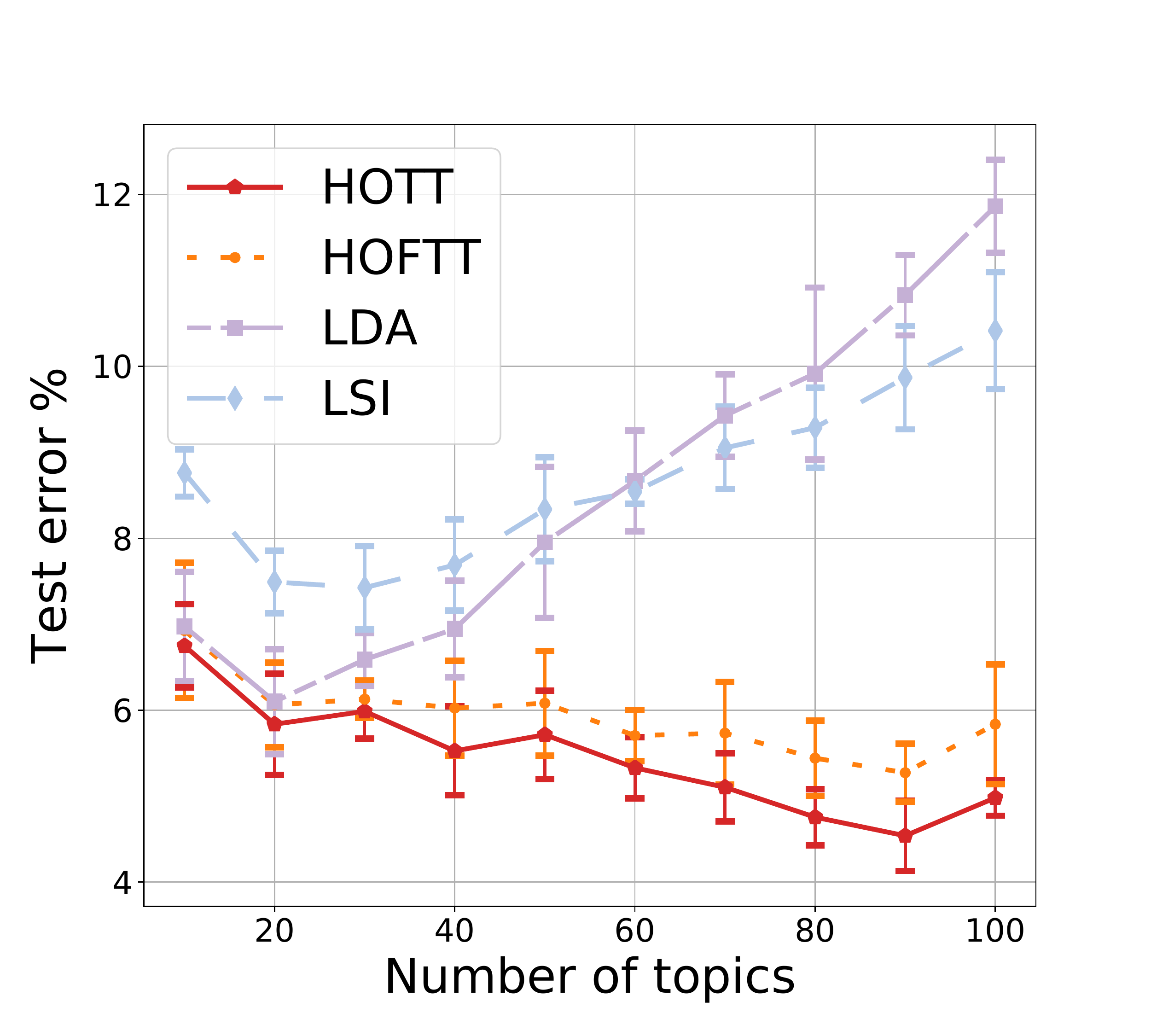}
\vspace{-.25in}
\caption{Topic number sensitivity on \textsc{classic}}
\label{fig:classic-K}
\end{subfigure}
\begin{subfigure}{.33\textwidth}
  \centering
  \captionsetup{justification=centering}
\vspace{-.14in}
\includegraphics[width=\linewidth]{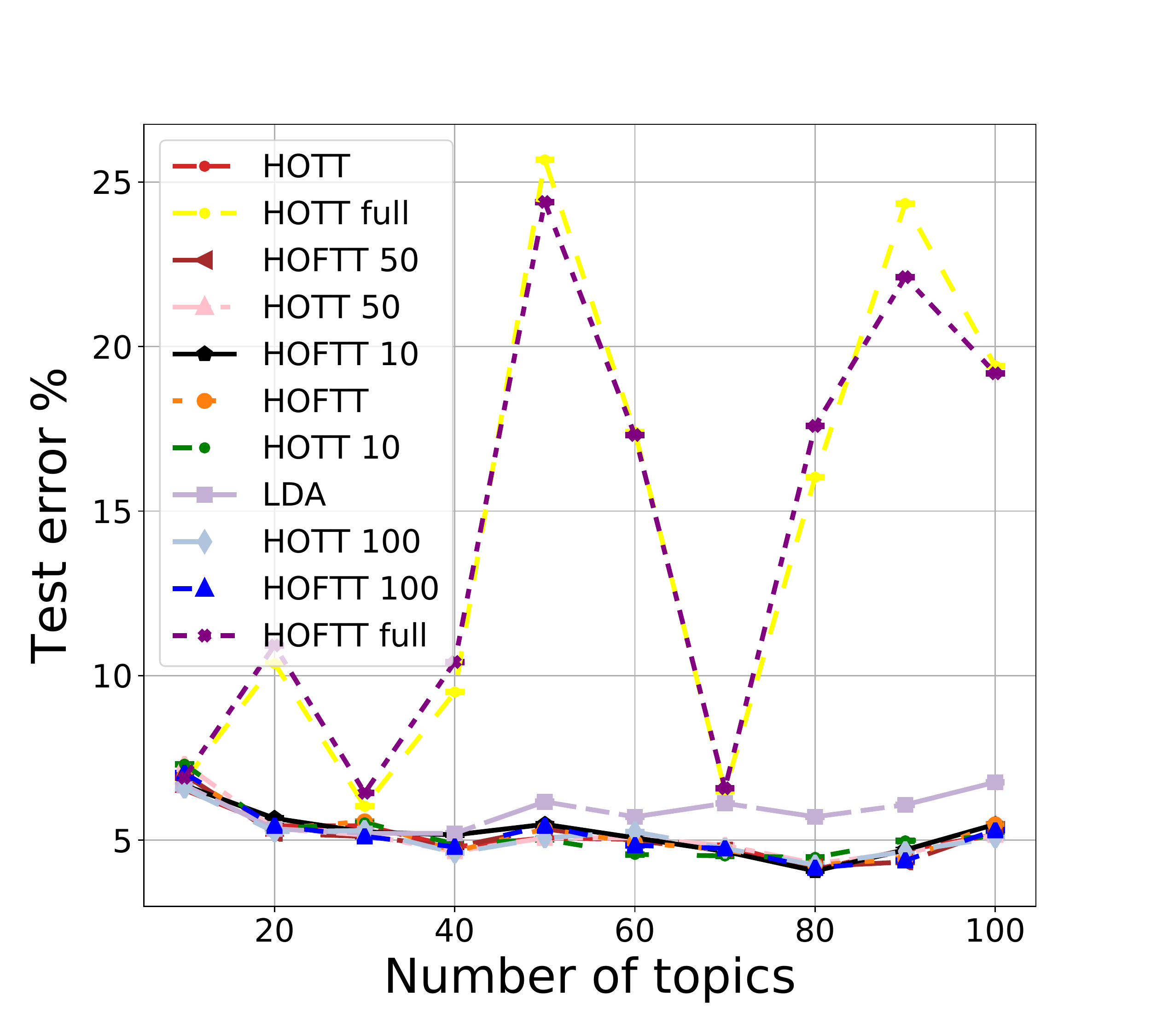}
\vspace{-.25in}
\caption{Topic truncation sensitivity on \textsc{reuters}}
\label{fig:r8-truncate}
\end{subfigure}
\caption{Sensitivity analysis: embedding, topic number and topic truncation}
\label{fig:sensitivity}
\end{figure*}

\textbf{Sensitivity to word embeddings.} 
We train \textit{word2vec} \citep{mikolov2013distributed} 200-dimensional embeddings on \textsc{Reuters} and compare relevant methods with our default embedding (i.e., \textit{GloVe}) and newly-trained \textit{word2vec} embeddings. According to \citet{mikolov2013distributed}, word embedding quality largely depends on data \emph{quantity} rather than quality; hence we expect the performance to degrade. In Fig.\ \ref{fig:r8-embed}, RWMD and WMD truncated performances drop as expected, but HOTT and HOFTT remain stable; this behavior is likely due to the embedding-independent topic structure taken into consideration.

\textbf{Number of LDA topics.} 
In our experiments, we set $\abs{T}=70$. When the $\abs{T}$ increases, LDA resembles the nBOW representation; correspondingly, HOTT approaches the WMD. The difference, however, is that nBOW is a weaker baseline, while WMD is powerful document distance. Using the \textsc{classic} dataset, in Fig.\ \ref{fig:classic-K} we demonstrate that LDA (and LSI) may degrade with too many topics, while HOTT and HOFTT are robust to topic overparameterization. In this example, better performance of HOTT over HOFTT is likely due relatively short documents of the {\sc{classic}} dataset.

While we have shown that HOTT is not sensitive to the choice of the number of topics, it is also possible to eliminate this parameter by using LDA inference algorithms that learn number of topics \citep{yurochkin2017conic} or adopting Bayesian nonparametric topic modes and corresponding inference schemes \citep{teh2006hierarchical, wang2011online, bryant2012truly}.

\textbf{Topic truncation.} 
Fig.\ \ref{fig:r8-truncate} demonstrates $k$-NN classification performance on the \textsc{reuters} dataset with varying topic truncation: top 10, 20 (HOTT and HOFTT), 50, 100 words and no truncation (HOTT full and HOFTT full); LDA performance is given for reference. Varying the truncation level does not affect the results significantly, however no truncation results in unstable performance.

\subsection{t-SNE metric visualization}

\begin{wrapfigure}[7]{r}{.4\linewidth}
\centering\vspace{-.2in}
\includegraphics[width=\linewidth]{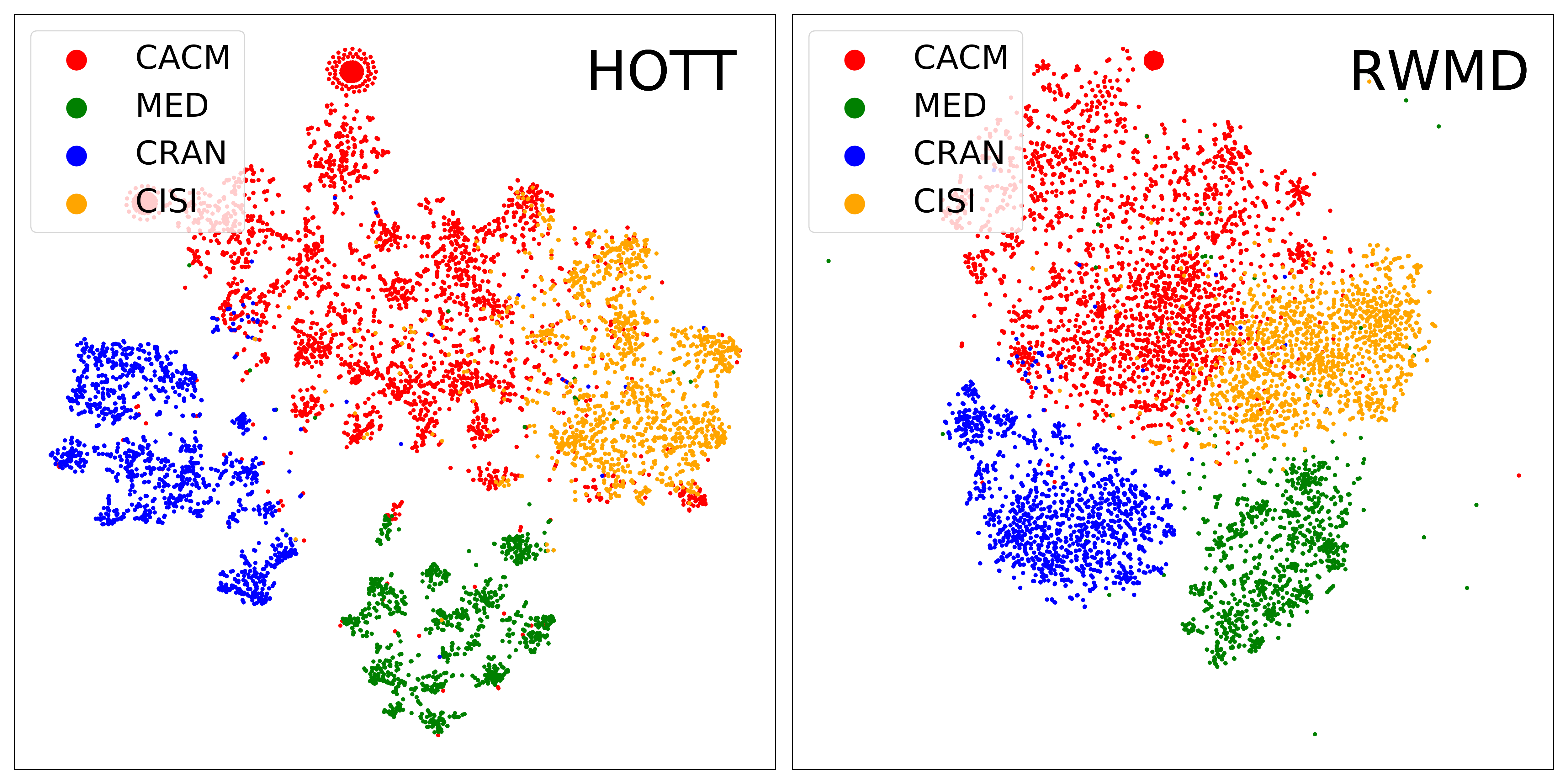}
\vspace{-.22in}
\caption{t-SNE on \textsc{classic}}\vspace{-.15in}
\label{fig:tSNE}
\end{wrapfigure}
Visualizing metrics as point clouds provides useful qualitative information for human users. Unlike $k$-NN classification, most methods for this task require long-range distances and a full metric. Here, we use t-SNE \citep{vanDerMaaten2008Visualizing} to visualize HOTT and RWMD on the \textsc{classic} dataset in Fig.\ \ref{fig:tSNE}. HOTT appears to more accurately separate the labeled points (color-coded). Appendix Fig. \ref{fig:supp_tsne} shows additional t-SNE results.

\subsection{Supervised link prediction}

We next evaluate HOTT in a different prediction task: supervised link prediction on graphs defined on text domains, here \emph{citation networks}. The specific task we address is the Kaggle challenge of Link Prediction TU.\footnote{\url{www.kaggle.com/c/link-prediction-tu}}
In this challenge, a citation network is given as an undirected graph, where nodes are research papers and (undirected) edges represent citations. From this 
graph, edges have been removed at random.  The task is to reconstruct the full network.
%
%
The dataset contains $27770$ papers (nodes).   The training and testing  sets consist of $615512$ and $32648$ node pairs (edges) respectively.  
For each paper, the available data only includes publication year, title, authors, and abstract.


\begin{table}[t]
 \caption{Link prediction: using distance (rows) for node-pair representations (cols).}\label{table:link_rediction}
\vspace{.05in}
\begin{center}
\begin{tabular}{r@{\quad}c@{\quad}c@{\quad}c@{\quad}c@{\quad}c} 
 \toprule
Distance  & & &F1 Score \\
\cmidrule[0.4pt](r{0.125em}){1-1}%
\cmidrule[0.4pt](lr{0.125em}){2-6}
    &$\Phi_0$  &  $\Phi_1$ & $\Phi_2$& $\Phi_3$ & $\Phi_4$  \\
\cmidrule[0.4pt](lr{0.125em}){2-6}%
%
   HOFTT& \bf{73.22}&\bf{76.27}&\bf{76.62}&\bf{78.85}&\bf{83.37}\\
    HOTT& 73.19&76.03&76.24&78.64&83.25\\
    RWMD&71.60&74.90&75.20&77.16&82.92\\
    WMD-T20&67.22 &63.38&65.20&70.38&81.84\\
    None& --- &61.13 &64.27&67.72 & 81.68\\
\bottomrule
 \end{tabular}
\end{center}
 \end{table}

To study the effectiveness of a distance-based model with HOTT for link prediction, we train a linear SVM classifier over the feature set $\Phi$, which includes the distance between the two abstracts $\phi_{dist}$ computed via one of  
\{HOFT, HOTT, RWMD, WMD-T20\}. For completeness, we also examine excluding the distance totally. 
We incrementally grow the feature sets $\Phi$ as: $\Phi_0 = \{\phi_{dist} \}$, $\Phi_1 =  \{ \phi_{dist}\} \cup  \{\phi_1\}$,
$\Phi_n = \{ \phi_{dist}\} \cup \{\phi_1, \dots, \phi_n\}  $ where $\phi_1$ is
the number of common words in the titles, $\phi_2$ the number of common authors,  and  $\phi_3$ and $\phi_4$ the signed  and absolute  difference between the publication years. 

Table \ref{table:link_rediction} presents the results; 
evaluation is based on the F1-Score. 
Consistently, HOFTT and HOTT  are more effective than   RWMD and WMD-T20 in all tests, and not using  any of the distances
consistently degrades the performance.

\section{Conclusion}
We have proposed a hierarchical method for comparing natural language documents that leverages optimal transport, topic modeling, and word embeddings. Specifically, word embeddings provide global semantic language information, while LDA topic models provide corpus-specific topics and topic distributions. Empirically these combine to give superior performance on various metric-based tasks.
We hypothesize that modeling documents by their representative topics is better for highlighting differences despite the loss in resolution. HOTT appears to capture differences in the same way a person asked to compare two documents would: by breaking down each document into easy to understand concepts, and then comparing the concepts.


There are many avenues for future work. From a theoretical perspective, our use of a nested Wasserstein metric suggests further analysis of this hierarchical transport space. Insight gained in this direction may reveal the learning capacity of our method and inspire faster or more accurate algorithms. 
%
%
From a computational perspective, our approach currently combines word embeddings, topic models and OT, but these are all trained separately. End-to-end training that efficiently optimizes these three components jointly would likely improve performance and facilitate analysis of our algorithm as a unified approach to document comparison.


Finally, from an empirical perspective, the performance improvements we observe stem directly from a reduction in the size of the transport problem. Investigation of larger corpora with longer documents, and applications requiring the full set of pairwise distances are now feasible. We also can consider applications to modeling of images or 3D data.  


\paragraph*{Acknowledgements.} 
J.\ Solomon acknowledges the generous support of Army Research Office grant W911NF1710068, Air Force Office of Scientific Research award FA9550-19-1-031, of National Science Foundation grant IIS-1838071, from an Amazon Research Award, from the MIT-IBM Watson AI Laboratory, from the Toyota-CSAIL Joint Research Center, from the QCRI--CSAIL Computer Science Research Program, and from a gift from Adobe Systems. Any opinions, findings, and conclusions or recommendations expressed in this material are those of the authors and do not necessarily reflect the views of these organizations.

\clearpage
\bibliography{MY_ref,transport_refs}

\begin{thebibliography}{38}
\providecommand{\natexlab}[1]{#1}
\providecommand{\url}[1]{\texttt{#1}}
\expandafter\ifx\csname urlstyle\endcsname\relax
  \providecommand{\doi}[1]{doi: #1}\else
  \providecommand{\doi}{doi: \begingroup \urlstyle{rm}\Url}\fi

\bibitem[Arora et~al.(2017)Arora, Liang, and Ma]{arora2016simple}
Arora, S., Liang, Y., and Ma, T.
\newblock A simple but tough-to-beat baseline for sentence embeddings.
\newblock In \emph{International Conference on Learning Representations}, 2017.

\bibitem[Atasu \& Mittelholzer(2019)Atasu and Mittelholzer]{atasu2019linear}
Atasu, K. and Mittelholzer, T.
\newblock Linear-complexity data-parallel earth mover’s distance
  approximations.
\newblock In \emph{International Conference on Machine Learning}, pp.\
  364--373, 2019.

\bibitem[Blei et~al.(2003)Blei, Ng, and Jordan]{blei2003latent}
Blei, D.~M., Ng, A.~Y., and Jordan, M.~I.
\newblock Latent {D}irichlet {A}llocation.
\newblock \emph{Journal of Machine Learning Research}, 3:\penalty0 993--1022,
  March 2003.

\bibitem[Broderick et~al.(2013)Broderick, Boyd, Wibisono, Wilson, and
  Jordan]{broderick2013streaming}
Broderick, T., Boyd, N., Wibisono, A., Wilson, A.~C., and Jordan, M.~I.
\newblock Streaming variational {B}ayes.
\newblock In \emph{Advances in Neural Information Processing Systems}, pp.\
  1727--1735, 2013.

\bibitem[Bryant \& Sudderth(2012)Bryant and Sudderth]{bryant2012truly}
Bryant, M. and Sudderth, E.~B.
\newblock Truly nonparametric online variational inference for hierarchical
  {D}irichlet processes.
\newblock In \emph{Advances in Neural Information Processing Systems}, pp.\
  2699--2707, 2012.

\bibitem[Cachopo et~al.(2007)]{cachopo2007improving}
Cachopo, A. M. d. J.~C. et~al.
\newblock Improving methods for single-label text categorization.
\newblock \emph{Instituto Superior T{\'e}cnico, Portugal}, 2007.

\bibitem[Cuturi(2013)]{cuturi2013sinkhorn}
Cuturi, M.
\newblock Sinkhorn distances: Lightspeed computation of optimal transport.
\newblock In \emph{Advances in Neural Information Processing Systems}, pp.\
  2292--2300, 2013.

\bibitem[Deerwester et~al.(1990)Deerwester, Dumais, Furnas, Landauer, and
  Harshman]{deerwester1990indexing}
Deerwester, S., Dumais, S.~T., Furnas, G.~W., Landauer, T.~K., and Harshman, R.
\newblock Indexing by latent semantic analysis.
\newblock \emph{Journal of the American Society for Information Science},
  41\penalty0 (6):\penalty0 391, Sep 01 1990.

\bibitem[Frakes \& Baeza-Yates(1992)Frakes and
  Baeza-Yates]{frakes1992information}
Frakes, W.~B. and Baeza-Yates, R.
\newblock \emph{Information retrieval: Data structures \& algorithms}, volume
  331.
\newblock prentice Hall Englewood Cliffs, NJ, 1992.

\bibitem[Griffiths \& Steyvers(2004)Griffiths and
  Steyvers]{griffiths2004finding}
Griffiths, T.~L. and Steyvers, M.
\newblock Finding scientific topics.
\newblock \emph{PNAS}, 101\penalty0 (suppl. 1):\penalty0 5228--5235, 2004.

\bibitem[Gurobi~Optimization(2018)]{gurobi}
Gurobi~Optimization, L.
\newblock Gurobi optimizer reference manual, 2018.
\newblock URL \url{http://www.gurobi.com}.

\bibitem[Hoffman et~al.(2013)Hoffman, Blei, Wang, and
  Paisley]{hoffman2013stochastic}
Hoffman, M.~D., Blei, D.~M., Wang, C., and Paisley, J.
\newblock Stochastic variational inference.
\newblock \emph{Journal of Machine Learning Research}, 14\penalty0
  (1):\penalty0 1303--1347, May 2013.

\bibitem[Huang et~al.(2016)Huang, Guo, Kusner, Sun, Sha, and
  Weinberger]{huang2016supervised}
Huang, G., Guo, C., Kusner, M.~J., Sun, Y., Sha, F., and Weinberger, K.~Q.
\newblock Supervised word mover's distance.
\newblock In \emph{Advances in Neural Information Processing Systems}, pp.\
  4862--4870, 2016.

\bibitem[Kuhn(1955)]{kuhn1955hungarian}
Kuhn, H.~W.
\newblock The {H}ungarian method for the assignment problem.
\newblock \emph{Naval Research Logistics (NRL)}, 2\penalty0 (1-2):\penalty0
  83--97, 1955.

\bibitem[Kusner et~al.(2015)Kusner, Sun, Kolkin, and
  Weinberger]{kusner2015word}
Kusner, M., Sun, Y., Kolkin, N., and Weinberger, K.
\newblock From word embeddings to document distances.
\newblock In \emph{International Conference on Machine Learning}, pp.\
  957--966, 2015.

\bibitem[Luhn(1957)]{luhn1957statistical}
Luhn, H.~P.
\newblock A statistical approach to mechanized encoding and searching of
  literary information.
\newblock \emph{IBM Journal of Research and Development}, 1\penalty0
  (4):\penalty0 309--317, 1957.

\bibitem[Mikolov et~al.(2013)Mikolov, Sutskever, Chen, Corrado, and
  Dean]{mikolov2013distributed}
Mikolov, T., Sutskever, I., Chen, K., Corrado, G.~S., and Dean, J.
\newblock Distributed representations of words and phrases and their
  compositionality.
\newblock In \emph{Advances in neural information processing systems}, pp.\
  3111--3119, 2013.

\bibitem[Newman et~al.(2010)Newman, Lau, Grieser, and
  Baldwin]{newman2010automatic}
Newman, D., Lau, J.~H., Grieser, K., and Baldwin, T.
\newblock Automatic evaluation of topic coherence.
\newblock In \emph{Human Language Technologies: The 2010 Annual Conference of
  the North American Chapter of the Association for Computational Linguistics},
  pp.\  100--108. Association for Computational Linguistics, 2010.

\bibitem[Otto \& Villani(2000)Otto and Villani]{otto2000generalization}
Otto, F. and Villani, C.
\newblock Generalization of an inequality by talagrand and links with the
  logarithmic sobolev inequality.
\newblock \emph{Journal of Functional Analysis}, 173\penalty0 (2):\penalty0
  361--400, 2000.

\bibitem[Pedregosa et~al.(2011)Pedregosa, Varoquaux, Gramfort, Michel, Thirion,
  Grisel, Blondel, Prettenhofer, Weiss, Dubourg, et~al.]{pedregosa2011scikit}
Pedregosa, F., Varoquaux, G., Gramfort, A., Michel, V., Thirion, B., Grisel,
  O., Blondel, M., Prettenhofer, P., Weiss, R., Dubourg, V., et~al.
\newblock Scikit-learn: {M}achine learning in {P}ython.
\newblock \emph{Journal of Machine Learning Research}, 12\penalty0
  (Oct):\penalty0 2825--2830, 2011.

\bibitem[Pennington et~al.(2014)Pennington, Socher, and
  Manning]{pennington2014glove}
Pennington, J., Socher, R., and Manning, C.~D.
\newblock Glove: Global vectors for word representation.
\newblock In \emph{Empirical Methods in Natural Language Processing (EMNLP)},
  pp.\  1532--1543, 2014.

\bibitem[Peyr\'e \& Cuturi(2018)Peyr\'e and Cuturi]{peyre2018computational}
Peyr\'e, G. and Cuturi, M.
\newblock \emph{Computational Optimal Transport}.
\newblock Submitted, 2018.

\bibitem[Sanders(2011)]{sanders2011sanders}
Sanders, N.~J.
\newblock Sanders-twitter sentiment corpus.
\newblock \emph{Sanders Analytics LLC}, 2011.

\bibitem[Santambrogio(2015)]{santambrogio_optimal_2015}
Santambrogio, F.
\newblock \emph{Optimal {{Transport}} for {{Applied Mathematicians}}},
  volume~87 of \emph{Progress in Nonlinear Differential Equations and Their
  Applications}.
\newblock {Springer International Publishing}, 2015.
\newblock ISBN 978-3-319-20827-5 978-3-319-20828-2.
\newblock \doi{10.1007/978-3-319-20828-2}.

\bibitem[Solomon(2018)]{solomon2018optimal}
Solomon, J.
\newblock \emph{Optimal Transport on Discrete Domains}.
\newblock AMS Short Course on Discrete Differential Geometry, 2018.

\bibitem[Sp\"arck~Jones(1972)]{sparck1972statistical}
Sp\"arck~Jones, K.
\newblock A statistical interpretation of term specificity and its application
  in retrieval.
\newblock \emph{Journal of Documentation}, 28\penalty0 (1):\penalty0 11--21,
  1972.

\bibitem[Teh et~al.(2006)Teh, Jordan, Beal, and Blei]{teh2006hierarchical}
Teh, Y.~W., Jordan, M.~I., Beal, M.~J., and Blei, D.~M.
\newblock Hierarchical {D}irichlet processes.
\newblock \emph{Journal of the American Statistical Association}, 101\penalty0
  (476), 2006.

\bibitem[van~der Maaten \& Hinton(2008)van~der Maaten and
  Hinton]{vanDerMaaten2008Visualizing}
van~der Maaten, L. and Hinton, G.
\newblock Visualizing data using {t-SNE}.
\newblock \emph{Journal of Machine Learning Research}, 9:\penalty0 2579--2605,
  2008.

\bibitem[Villani(2009)]{villani_optimal_2009}
Villani, C.
\newblock \emph{Optimal Transport: Old and New}.
\newblock Number 338 in Grundlehren der mathematischen Wissenschaften.
  {Springer}, Berlin, 2009.
\newblock ISBN 978-3-540-71049-3.
\newblock OCLC: ocn244421231.

\bibitem[Wan(2007)]{wan2007document}
Wan, X.
\newblock A novel document similarity measure based on earth mover’s
  distance.
\newblock \emph{Information Sciences}, 177\penalty0 (18):\penalty0 3718 --
  3730, 2007.
\newblock ISSN 0020-0255.
\newblock \doi{https://doi.org/10.1016/j.ins.2007.02.045}.

\bibitem[Wang et~al.(2011)Wang, Paisley, and Blei]{wang2011online}
Wang, C., Paisley, J., and Blei, D.
\newblock Online variational inference for the hierarchical {D}irichlet
  process.
\newblock In \emph{Proceedings of the 14th International Conference on
  Artificial Intelligence and Statistics}, pp.\  752--760, 2011.

\bibitem[Williamson et~al.(2010)Williamson, Wang, Heller, and
  Blei]{williamson2010ibp}
Williamson, S., Wang, C., Heller, K.~A., and Blei, D.~M.
\newblock The {IBP} compound {D}irichlet process and its application to focused
  topic modeling.
\newblock In \emph{Proceedings of the 27th International Conference on Machine
  Learning}, pp.\  1151--1158, 2010.

\bibitem[Wu et~al.(2018)Wu, Yen, Xu, Xu, Balakrishnan, Chen, Ravikumar, and
  Witbrock]{wu2018word}
Wu, L., Yen, I.~E., Xu, K., Xu, F., Balakrishnan, A., Chen, P.-Y., Ravikumar,
  P., and Witbrock, M.~J.
\newblock Word mover's embedding: From word2vec to document embedding.
\newblock \emph{Proceedings of the 2018 Conference on Empirical Methods in
  Natural Language Processing}, pp.\  4524--4534, 2018.

\bibitem[Wu \& Li(2017)Wu and Li]{wu2017topic}
Wu, X. and Li, H.
\newblock Topic mover's distance based document classification.
\newblock In \emph{Communication Technology (ICCT), 2017 IEEE 17th
  International Conference on}, pp.\  1998--2002. IEEE, 2017.

\bibitem[Xu et~al.(2018)Xu, Wang, Liu, and Carin]{xu2018distilled}
Xu, H., Wang, W., Liu, W., and Carin, L.
\newblock Distilled wasserstein learning for word embedding and topic modeling.
\newblock In \emph{Advances in Neural Information Processing Systems}, pp.\
  1716--1725, 2018.

\bibitem[Yurochkin \& Nguyen(2016)Yurochkin and Nguyen]{yurochkin2016geometric}
Yurochkin, M. and Nguyen, X.
\newblock Geometric {D}irichlet {M}eans {A}lgorithm for topic inference.
\newblock In \emph{Advances in Neural Information Processing Systems}, pp.\
  2505--2513, 2016.

\bibitem[Yurochkin et~al.(2017)Yurochkin, Guha, and Nguyen]{yurochkin2017conic}
Yurochkin, M., Guha, A., and Nguyen, X.
\newblock Conic {S}can-and-{C}over algorithms for nonparametric topic modeling.
\newblock In \emph{Advances in Neural Information Processing Systems}, pp.\
  3881--3890, 2017.

\bibitem[Yurochkin et~al.(2019)Yurochkin, Guha, Sun, and
  Nguyen]{yurochkin2019dirichlet}
Yurochkin, M., Guha, A., Sun, Y., and Nguyen, X.
\newblock Dirichlet simplex nest and geometric inference.
\newblock In \emph{International Conference on Machine Learning}, pp.\
  7262--7271, 2019.

\end{thebibliography}
\bibliographystyle{icml2019}

\clearpage
\appendix

\section{Metric properties}
\label{supp:metric}

$\HOTT$ is a metric in the lifted topic space since $W_p$ is a metric on distributions.

\begin{proof}
We can additionally prove that if we can exactly write $d^i = \sum_{k=1}^{|T|} \bar{d}_k^i t_k$ and if $t_i \neq t_j$ for $i \neq j$, then $\HOTT$ is a metric in document space. 

Positivity, symmetry, and the triangle inequality follow from properties of $W_2$. We prove that if $\HOTT(d^i, d^j) = 0$, then $d^i = d^j$. From the definition of $\HOTT$, 
\begin{align*}
    \HOTT(d^i, d^j) = W_2\left(\sum_{k=1}^{|T|} \bar{d}_k^i \delta_{t_k}, \sum_{l=1}^{|T|} \bar{d}_l^j \delta_{t_l}\right).
\end{align*}
If $\HOTT(d^i, d^j) = 0$, then if the transport plan is positive at $T_{k, l}$, it must hold that $W_p(t_k, t_l) = 0$. Since $W_p$ is a metric on probability distributions, this implies $t_k = t_l$. As we assumed that topics are distinct, and that documents are uniquely represented as linear combinations of topics we have $d^i = d^j$.
\end{proof}

\section{HOTT/WMD/RWMD relation}
\label{supp:rwmd_relation}

Following the discussion in Section \ref{sec:tmd} of the main text, we relate HOTT and RWMD to WMD empirically in terms of Mantel correlation and a Frobenius norm. 
The results are in Table \ref{table:dists}. While it is unsurprising that RWMD is more strongly correlated with WMD (HOTT is neither a lower nor an upper bound), we note that HOTT is on average a better approximation to WMD than RWMD. 

\begin{table}[h]
\centering
\small\addtolength{\tabcolsep}{0pt}
\caption{Relation between the metrics. For each dataset, we compute distance matrices using exact WMD, RWMD, and HOTT from a few randomly-selected documents. 
We report results of a Mantel correlation test between WMD/HOTT and WMD/RWMD and 
the difference between cost matrices under a Frobenius norm.}
\vspace{0.2in}
\label{table:dists}
\begin{tabular}{l rr rr}
\toprule
&
\multicolumn{2}{c}{Mantel} & 
\multicolumn{2}{c}{$l_2$} \\

\cmidrule[0.4pt](lr{0.125em}){2-3}%
\cmidrule[0.4pt](lr{0.125em}){4-5}%
Dataset & HOTT & RWMD & HOTT & RWMD\\
\midrule
\textsc{ohsumed} & 0.57 & 0.87 & 55 & 104\\
\textsc{20news} & 0.62 & 0.90 & 90 & 99\\
\textsc{amazon} & 0.49 & 0.84 & 70 & 65\\
\textsc{reuters} & 0.72 & 0.91 & 130 & 151\\
\textsc{bbcsport} & 0.76 & 0.92 & 28 & 90\\
\textsc{classic} & 0.43 & 0.89 & 157 & 69\\
\midrule
Avg & 0.60 & 0.89 & 88 & 96\\
\bottomrule
\end{tabular}
\end{table}

\section{Additional experimental results}
\label{supp:experiments}

In the main text, we used $W_1$ distance and did not do any vocabulary reduction, following the experimental setup of \citet{kusner2015word}. $W_2$ distance has intuitive geometric properties and is equipped with a variety of theoretical characterizations \citep{villani_optimal_2009}; one intuition for the difference between $W_1$ and $W_2$ comes from an analogy to the differences between $l_1$ and $l_2$ regularization. On the other hand, stemming is a common vocabulary reduction technique to improve quality of topic models. Stemming attempts to merge terms which differ only in their ending, i.e. ``cat'' and ``cats''. As stemming sometimes produces words not available in the \textit{GloVe} embeddings \citep{pennington2014glove}, to embed a stemmed word we take the average embeddings of the words mapped to it. We used \textit{SnowballStemmer} available from the \textit{nltk} Python package.

\begin{figure}[t]\centering
\includegraphics[width=\textwidth]{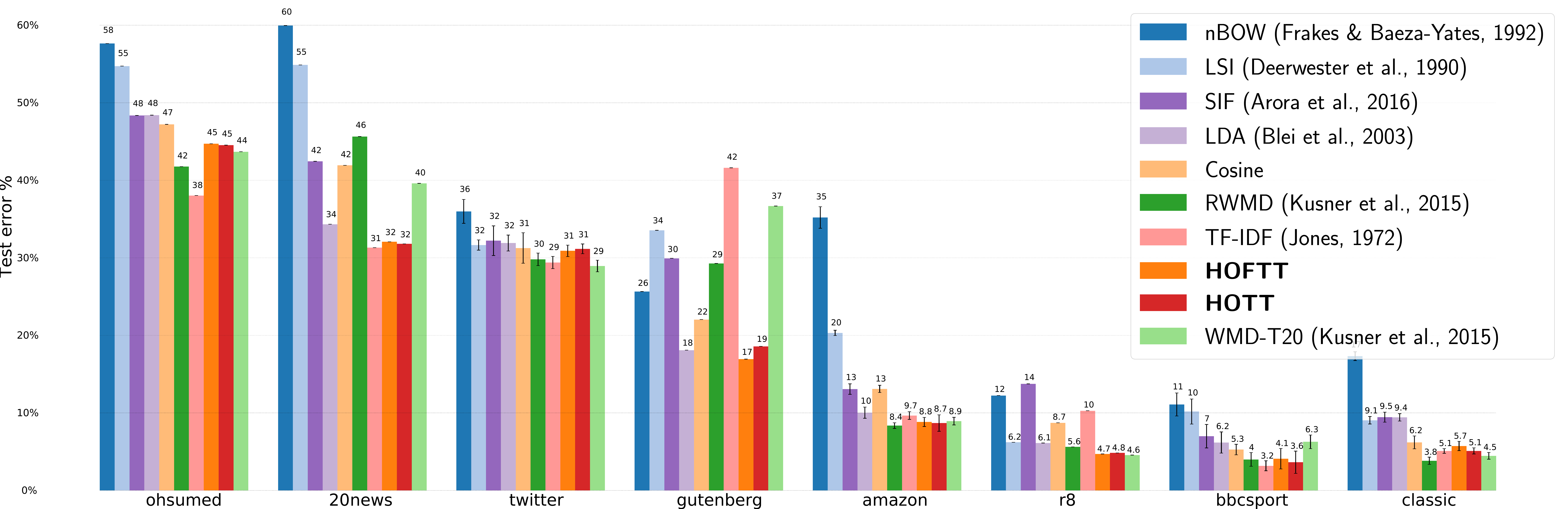}
\caption{$W_1$ and stemming: $k$-NN classification performance across datasets}
\vspace{-.2in}
\label{fig:w1_stem_kNN}
\end{figure}

\begin{figure}[t]\centering
\includegraphics[width=0.7\linewidth]{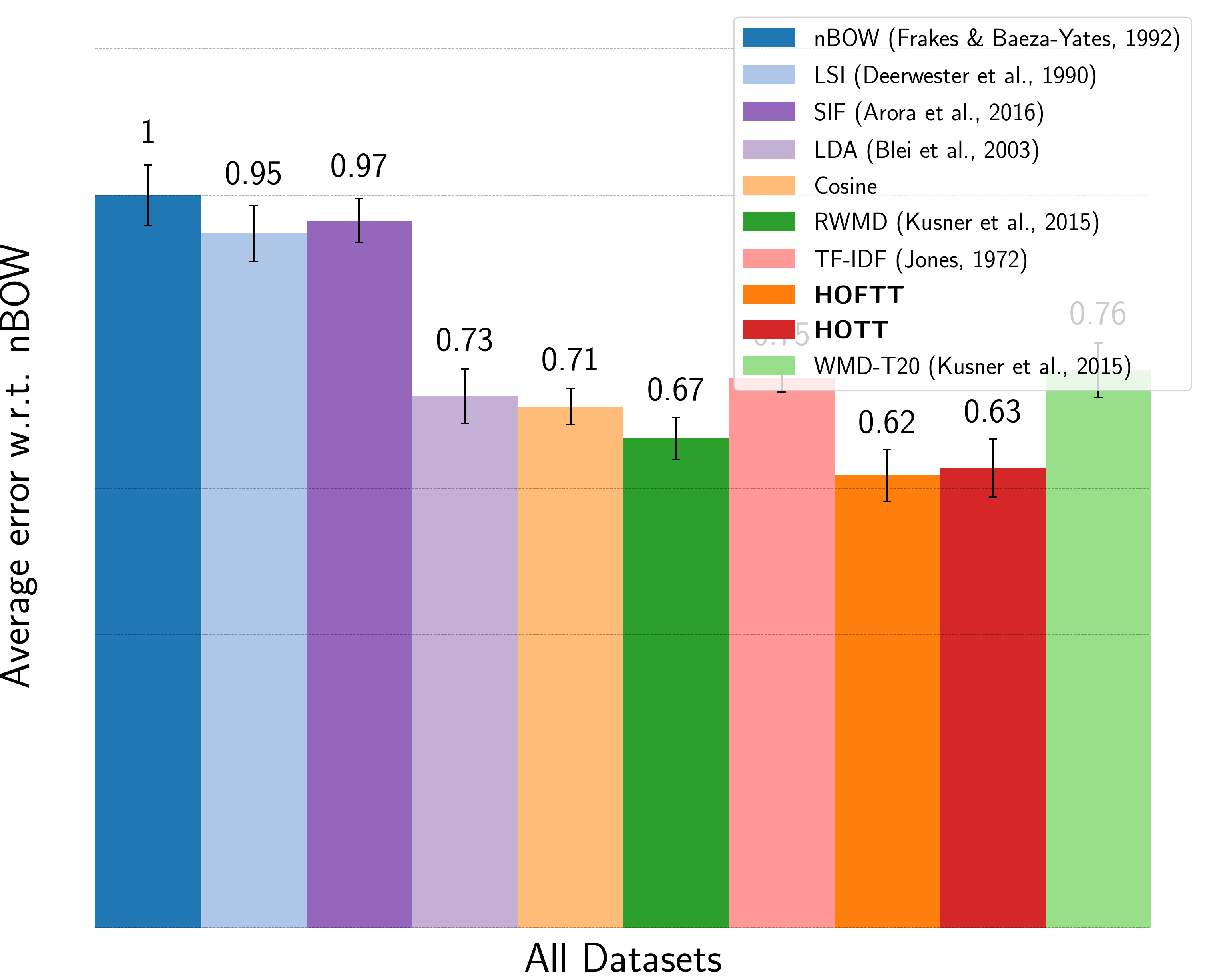}
\caption{$W_1$ and stemming: $k$-NN classification performance normalized by nBOW}\vspace{-.3in}
\label{fig:w1_stem_overallkNN}
\end{figure}

\begin{figure*}[t]\centering
\includegraphics[width=\textwidth]{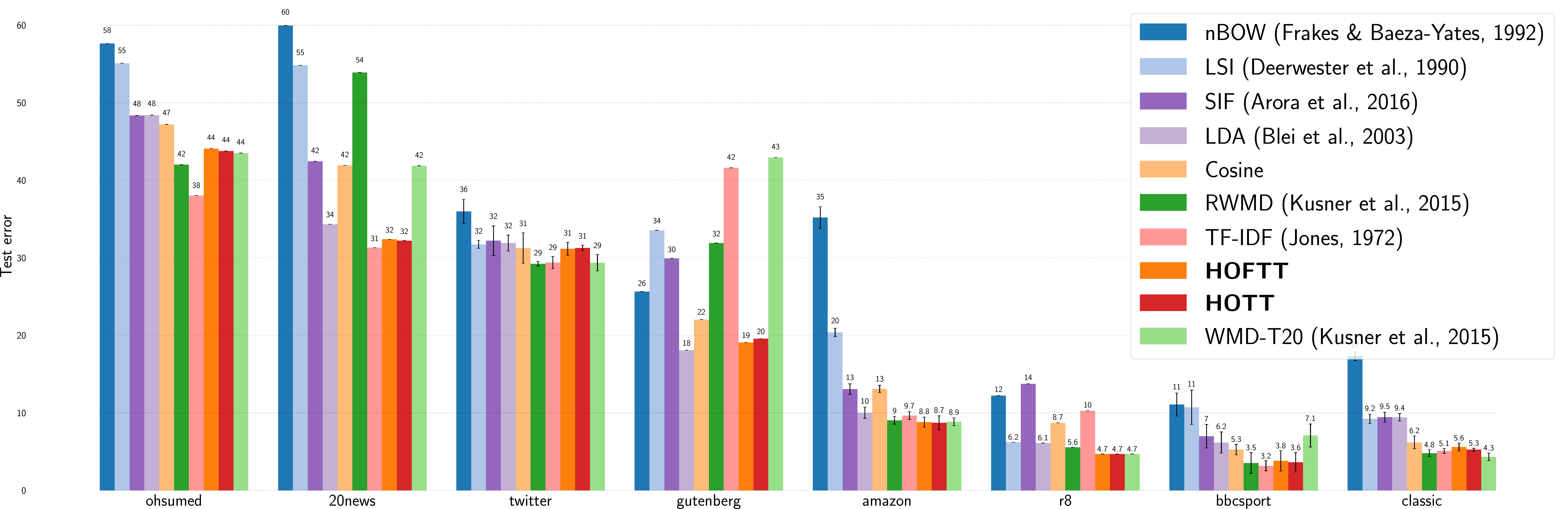}
\caption{$W_2$ without stemming: $k$-NN classification performance across datasets}
\vspace{-.2in}
\label{fig:w2_kNN}
\end{figure*}

\begin{figure}[t]\centering
\includegraphics[width=0.7\linewidth]{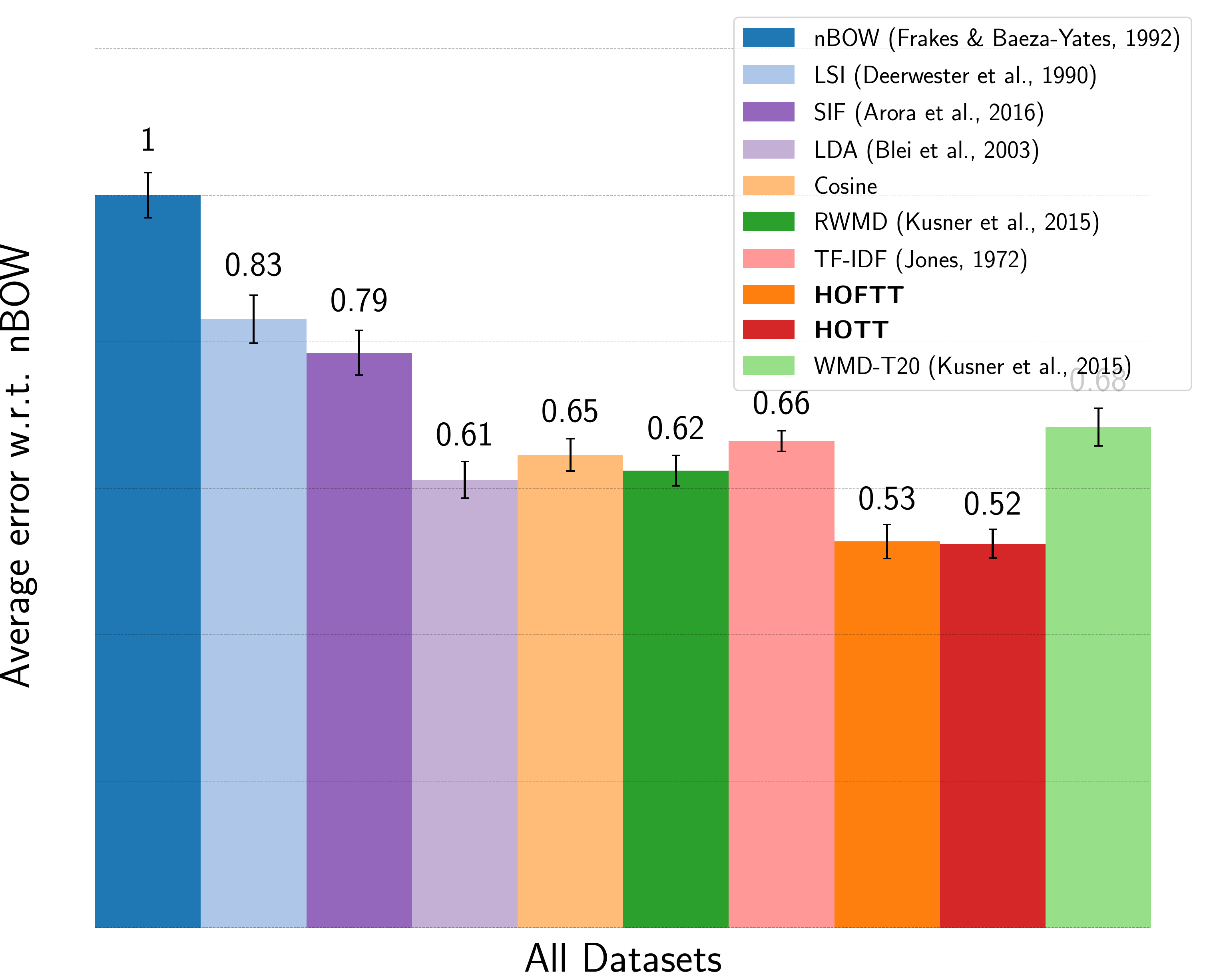}
\caption{$W_2$ without stemming: aggregated $k$-NN classification performance normalized by nBOW}\vspace{-.3in}
\label{fig:w2_overallkNN}
\end{figure}

\begin{figure*}[t]\centering
\includegraphics[width=\textwidth]{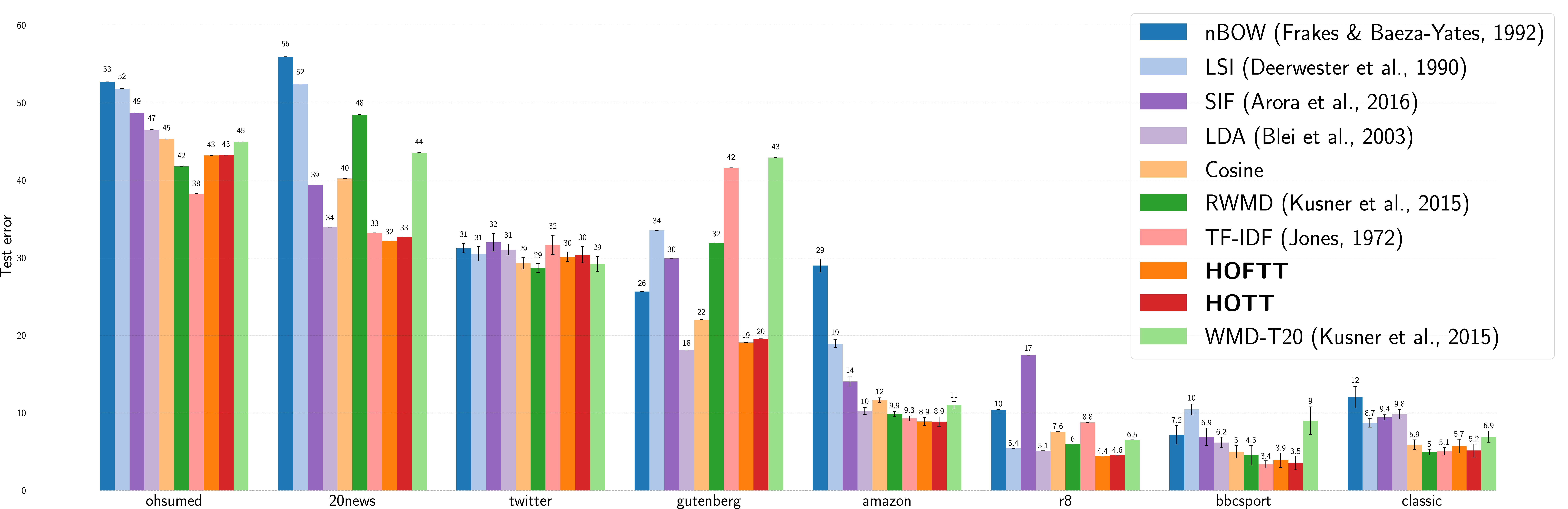}
\caption{$W_2$ and stemming: $k$-NN classification performance across datasets}
\vspace{-.2in}
\label{fig:w2_stem_kNN}
\end{figure*}

\begin{figure}[t]\centering
\includegraphics[width=0.7\linewidth]{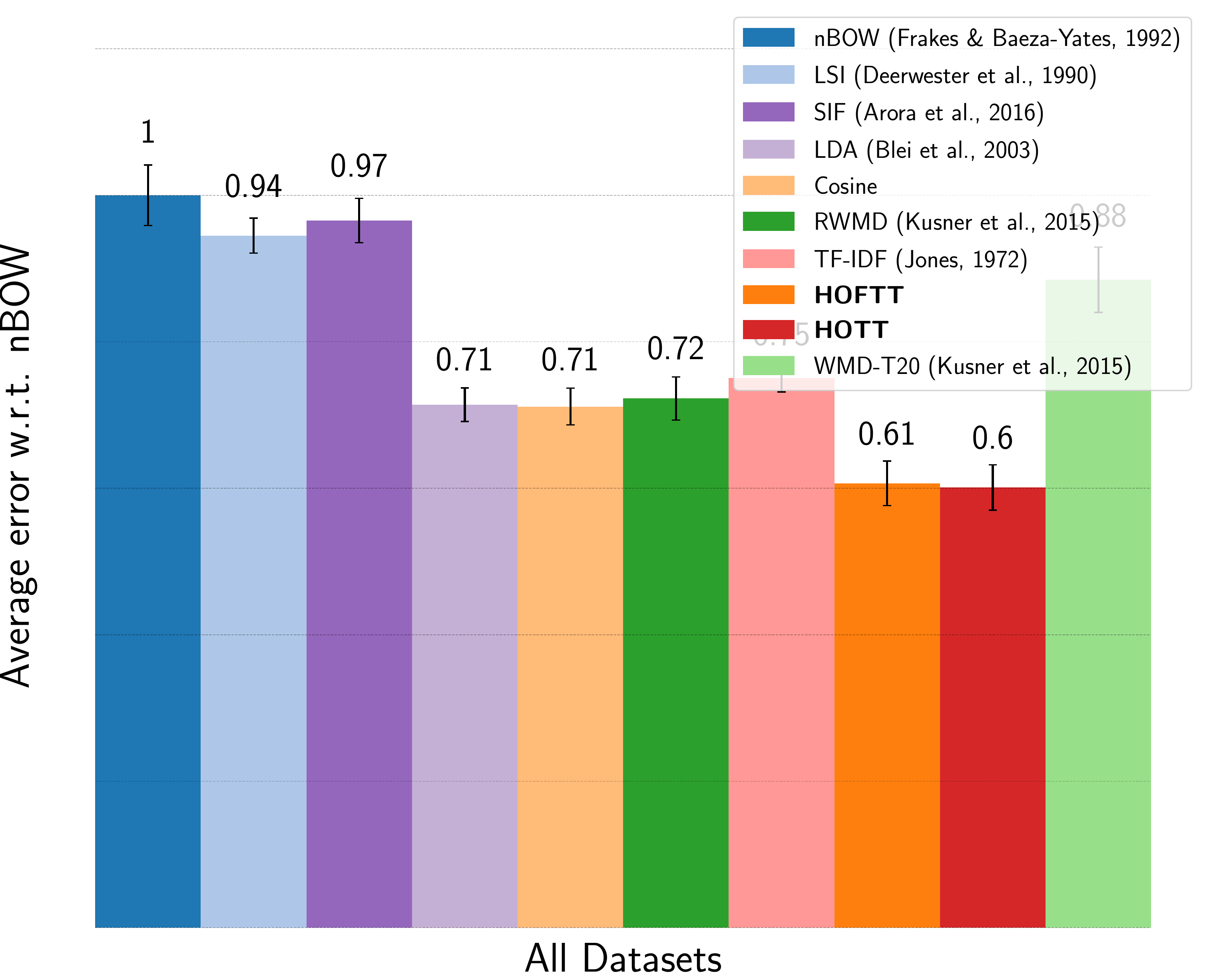}
\caption{$W_2$ and stemming: aggregated $k$-NN classification performance normalized by nBOW}\vspace{-.3in}
\label{fig:w2_stem_overallkNN}
\end{figure}

Figures \ref{fig:w1_stem_kNN} and \ref{fig:w1_stem_overallkNN} demonstrate results with $W_1$ and stemming; Figures \ref{fig:w2_kNN} and \ref{fig:w2_overallkNN} with $W_2$ and no stemming; Figures \ref{fig:w2_stem_kNN} and \ref{fig:w2_stem_overallkNN} with $W_2$ and stemming. In all settings HOTT and HOFTT are the best on average. Interestingly, using $W_2$ degrades performance of RWMD and WMD-T20, while our methods perform equally well with $W_1$ and $W_2$. Stemming tends to improve performance of nBOW, therefore aggregated results appear worse. Stemming also negatively effects RWMD and WMD-T20, while appears to have no effect on HOTT and HOFTT. For example, in the case of $W_2$ with stemming (Figures \ref{fig:w2_stem_kNN} and \ref{fig:w2_stem_overallkNN}), RWMD is no longer superior to baselines LDA \citep{blei2003latent} and Cosine, while our methods maintain good performance. We conclude that our methods are more robust to the choice of text processing techniques and specifics of the Wasserstein distance.

\begin{figure*}[ht]
\begin{subfigure}{.5\textwidth}
  \centering
  \captionsetup{justification=centering}
\includegraphics[width=\linewidth]{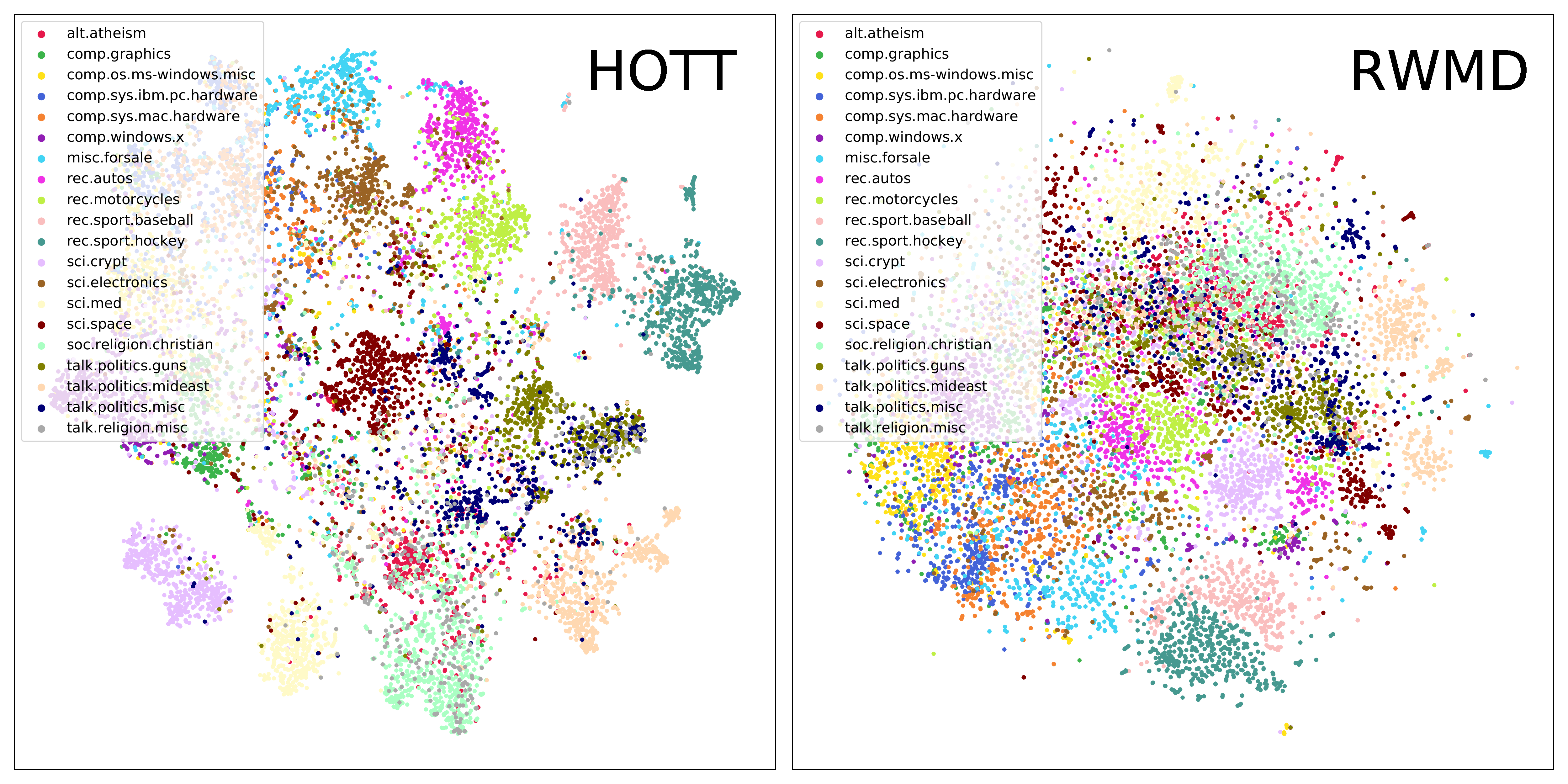}
\caption{\textsc{20news}}
\end{subfigure}
\begin{subfigure}{.5\textwidth}
  \centering
  \captionsetup{justification=centering}
\includegraphics[width=\linewidth]{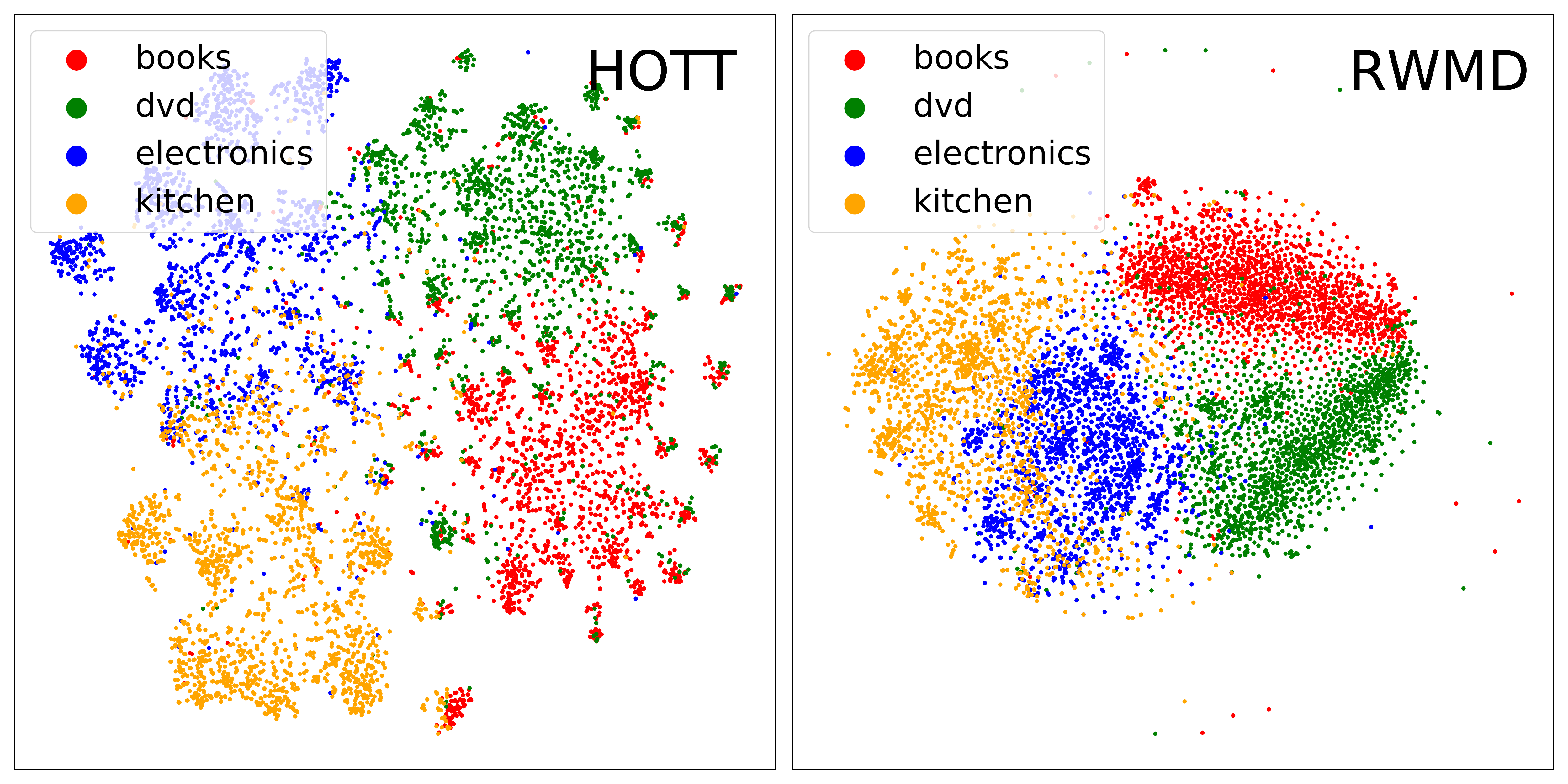}
\caption{\textsc{amazon}}
\end{subfigure}
\begin{subfigure}{.5\textwidth}
  \centering
  \captionsetup{justification=centering}
\includegraphics[width=\linewidth]{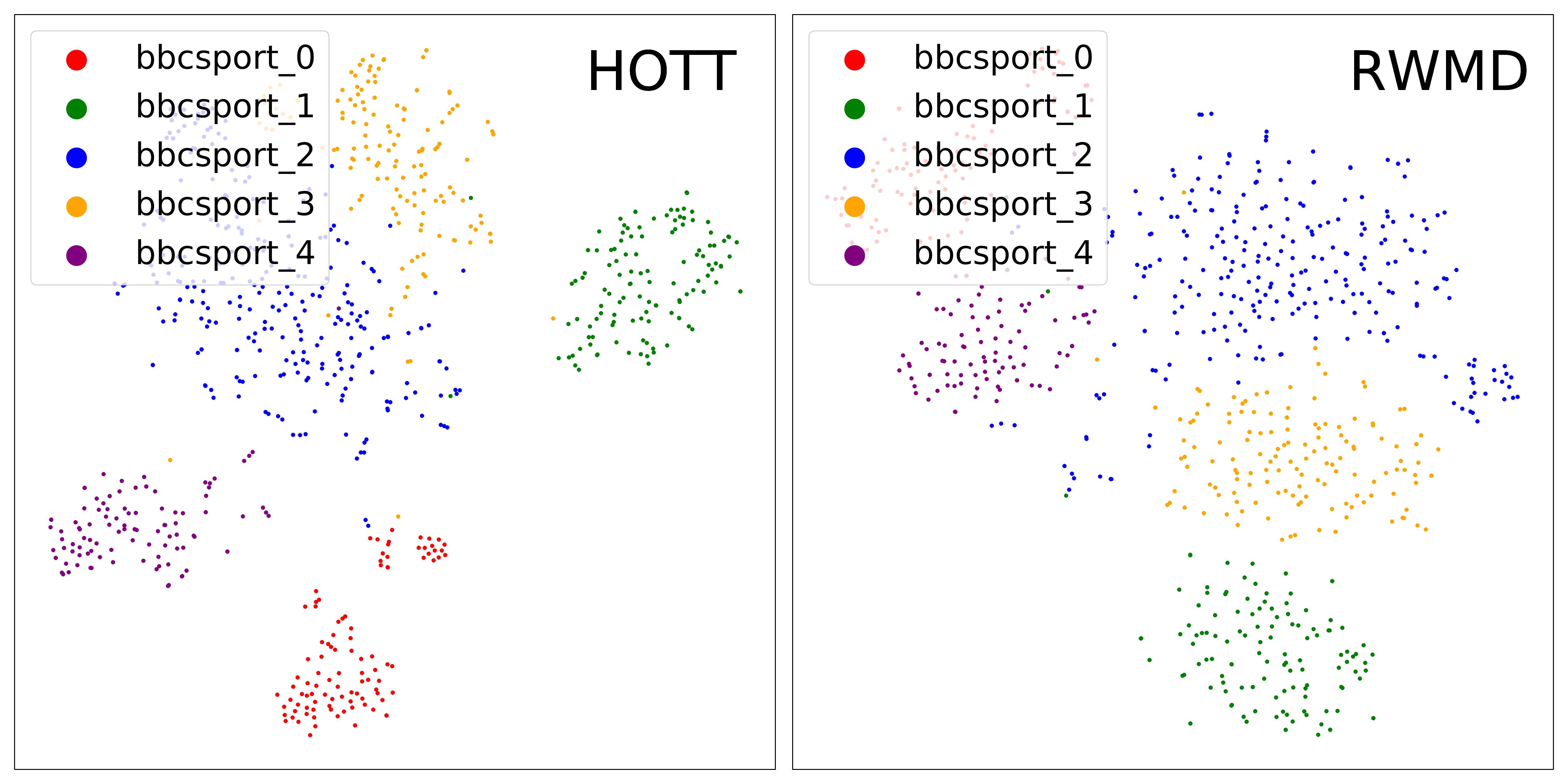}
\caption{\textsc{bbcsport}}
\end{subfigure}
\begin{subfigure}{.5\textwidth}
  \centering
  \captionsetup{justification=centering}
\includegraphics[width=\linewidth]{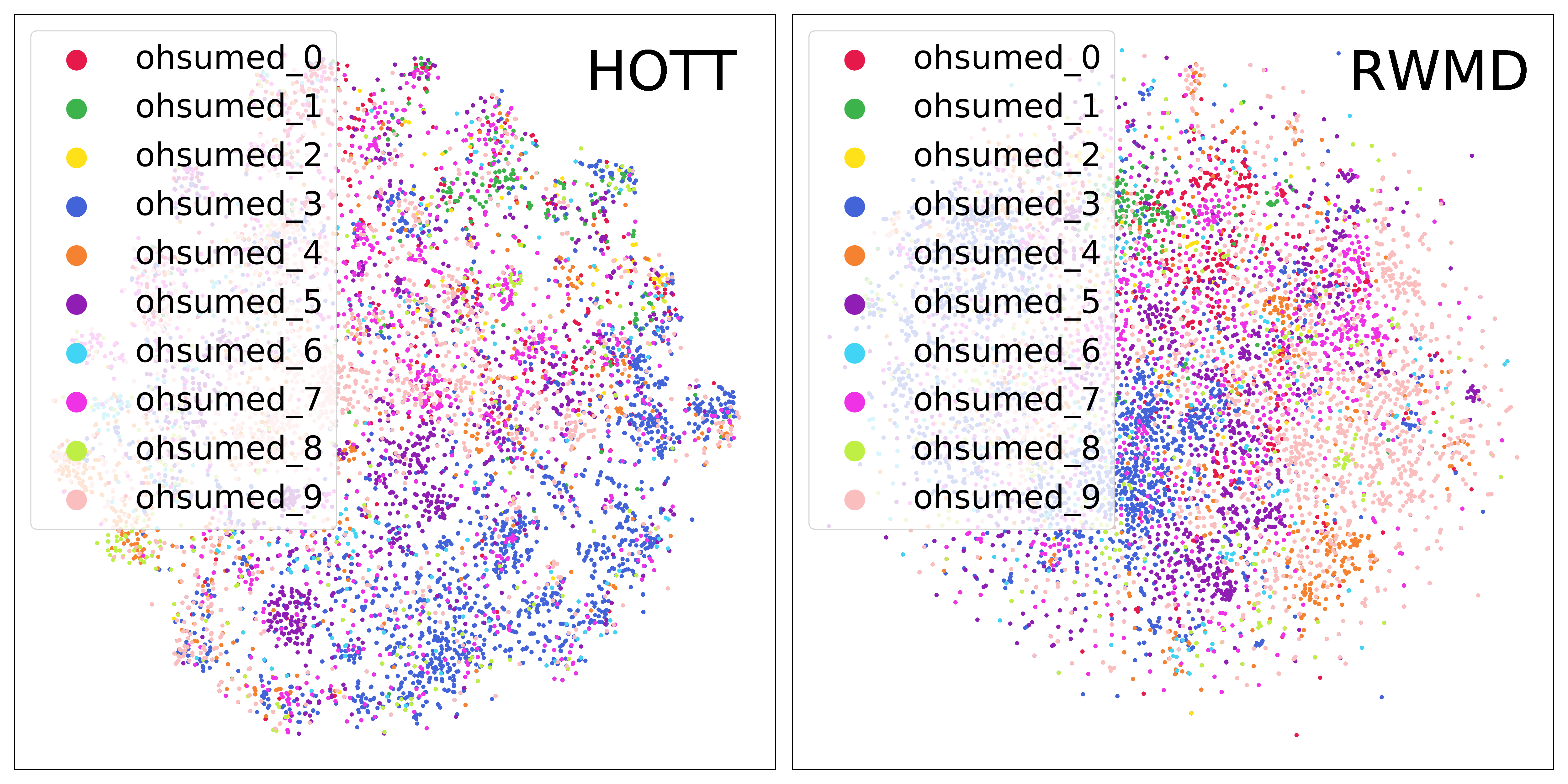}
\caption{\textsc{ohsumed}}
\end{subfigure}
\begin{subfigure}{.5\textwidth}
  \centering
  \captionsetup{justification=centering}
\includegraphics[width=\linewidth]{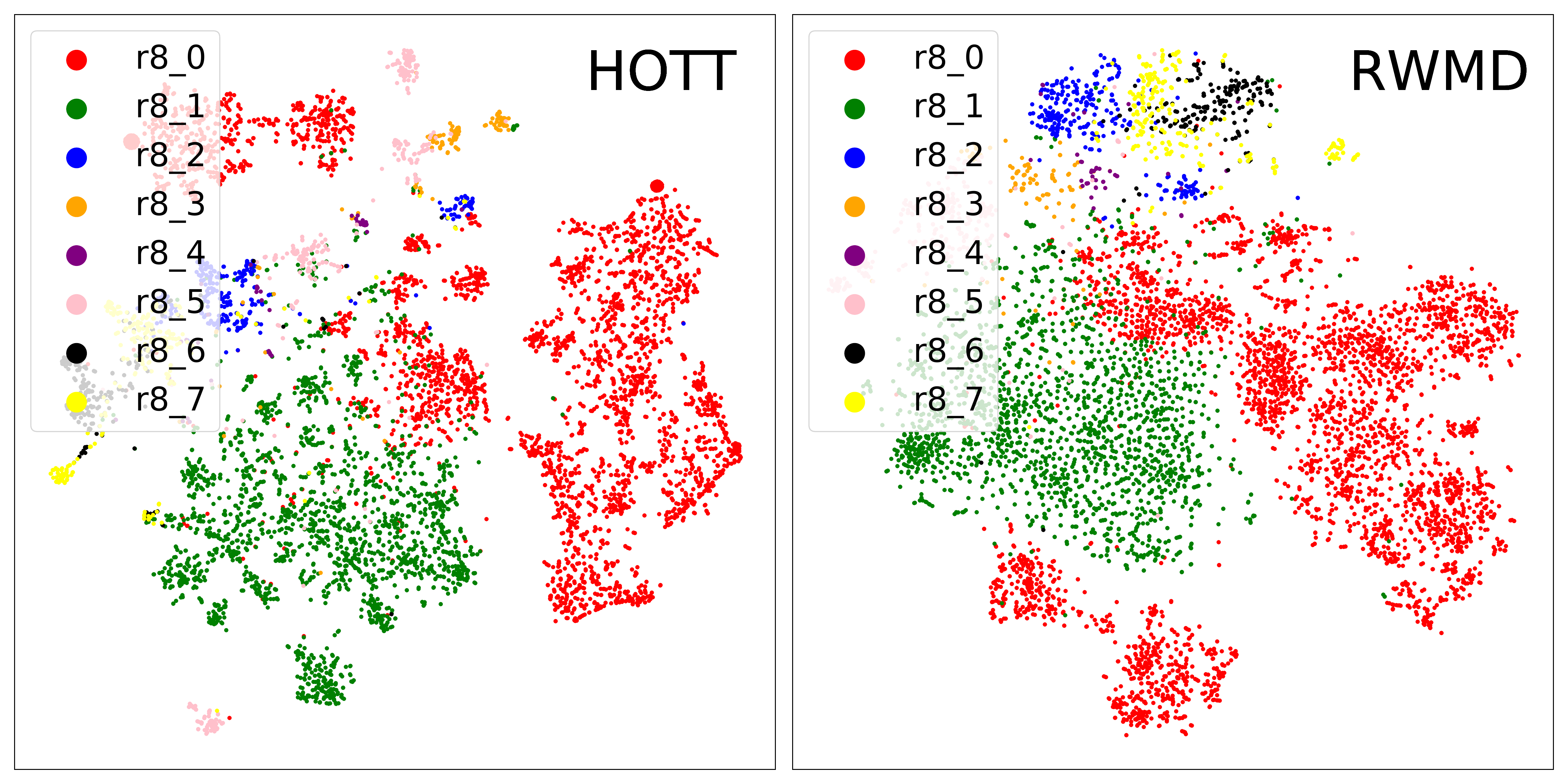}
\caption{\textsc{reuters}}
\end{subfigure}
\begin{subfigure}{.5\textwidth}
  \centering
  \captionsetup{justification=centering}
\includegraphics[width=\linewidth]{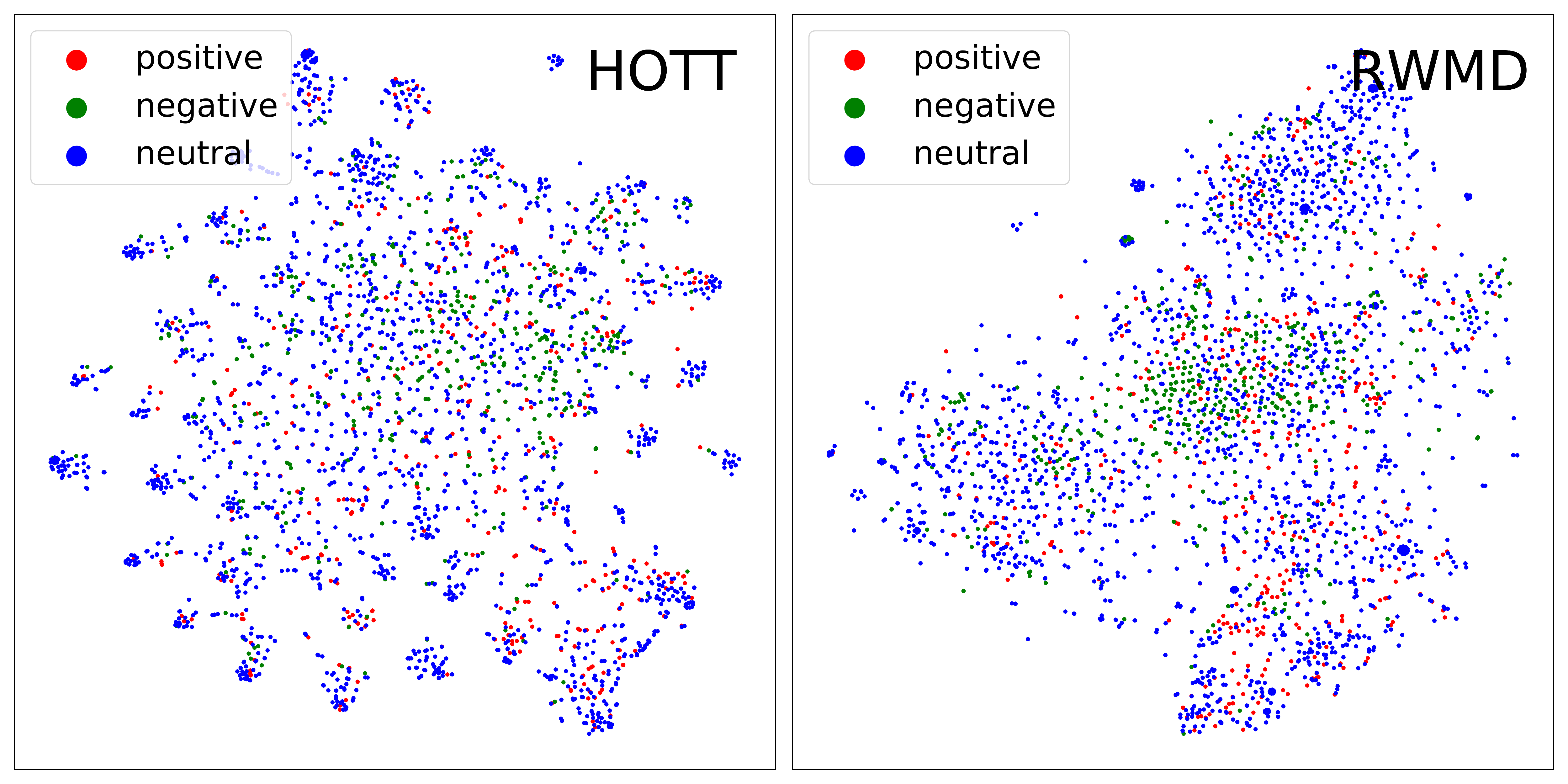}
\caption{\textsc{twitter}}
\end{subfigure}
\caption{These are the additional t-SNE results on all other datasets, except \textsc{gutenberg}, which is excluded due to its high number of classes (142). These images show that clusters based on our metric better align with the labels, corresponding to a better metric than RWMD. Both methods perform poorly on \textsc{twitter}, a difficult dataset for topic modelling.}
\label{fig:supp_tsne}
\end{figure*}

In Figure \ref{fig:supp_tsne} we present additional t-SNE \citep{vanDerMaaten2008Visualizing} visualization results.

\clearpage

\clearpage

\clearpage

\clearpage

\end{document}